\documentclass[10pt,twocolumn,letterpaper]{article}

\usepackage[pagenumbers]{cvpr} 

\usepackage[dvipsnames]{xcolor}

\usepackage{times}
\usepackage{epsfig}
\usepackage{graphicx}
\usepackage{amsmath}
\usepackage{amssymb}
\usepackage{booktabs}
\usepackage{caption}
\usepackage{adjustbox}
\usepackage{pifont}
\usepackage{tabu}
\usepackage{float}
\usepackage{balance}
\usepackage{multicol}
\usepackage{pgf-pie}
\usepackage{subcaption}
\usepackage{multirow,tabularx}
\usepackage[font=small,skip=5pt,belowskip=-9pt]{caption}

\definecolor{cvprblue}{rgb}{0.21,0.49,0.74}
\usepackage[pagebackref,breaklinks,colorlinks,citecolor=cvprblue]{hyperref}
\usepackage[capitalize]{cleveref}
\crefname{section}{Sec.}{Secs.}
\Crefname{section}{Section}{Sections}
\Crefname{table}{Table}{Tables}
\crefname{table}{Tab.}{Tabs.}

\usepackage{bm}

\DeclareMathOperator*{\argmin}{arg\,min}

\renewcommand{\paragraph}[1]{\vspace{0.5mm}\noindent\textbf{#1}\:}

\newcommand{\bbeta}{\boldsymbol{\beta}}

\newcommand{\btheta}{\boldsymbol{\theta}}
\newcommand{\bomega}{\boldsymbol{\omega}}

\newcommand{\bphi}{\boldsymbol{\phi}}

\newcommand{\cb}{\mathbf{c}}

\newcommand{\xb}{\mathbf{x}}
\newcommand{\yb}{\mathbf{y}}

\newcommand{\zz}{\mathbf{z}}

\newcommand{\hh}{\mathbf{h}}

\newcommand{\Xb}{\mathbf{X}}
\newcommand{\Yb}{\mathbf{Y}}
\newcommand{\Qb}{\mathbf{Q}}
\newcommand{\Hb}{\mathbf{H}}
\newcommand{\Pb}{\mathbf{P}}
\newcommand{\Vb}{\mathbf{V}}
\newcommand{\Sb}{\mathbf{S}}
\newcommand{\Fb}{\mathbf{F}}

\newcommand{\cmark}{\ding{51}}%
\newcommand{\xmark}{\ding{55}}%
\newcolumntype{R}[2]{%
    >{\adjustbox{angle=#1,lap=\width-(#2)}\bgroup}%
    l%
    <{\egroup}%
}
\newcommand*\rot{\multicolumn{1}{R{30}{1em}}}

\renewcommand{\paragraph}[1]{\vspace{0.5mm}\noindent\textbf{#1}\:}

\def\paperID{117} 
\def\confName{3DV\xspace}
\def\confYear{2024\xspace}

\title{\vspace{-8mm} {\vspace{-1mm} SPHEAR: Spherical Head Registration for Complete Statistical 3D Modeling}}

\author{Eduard Gabriel Bazavan, Andrei Zanfir, Thiemo Alldieck, Teodor Alexandru Szente, Mihai Zanfir \\ Cristian Sminchisescu
\\
\and 
{\bf Google Research}\\
{\tt\small \{egbazavan, andreiz, alldieck, teosz, mihaiz, sminchisescu\}@google.com} \\
}

\begin{document}
\twocolumn[{
\renewcommand\twocolumn[1][]{#1}
\maketitle
\thispagestyle{empty}
\vspace{-0.6cm}
\centering
\includegraphics[width=0.9\linewidth, trim={0.1cm, 0.1cm 0.1cm 0.1cm}, clip=true]{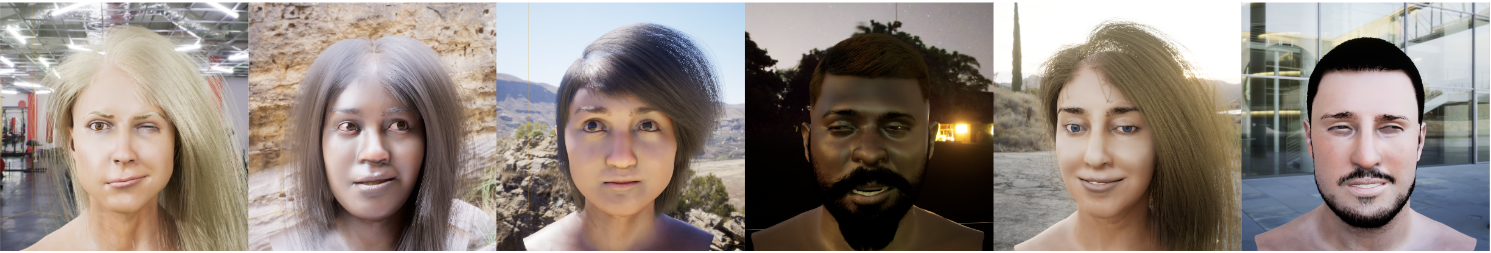} \\

\captionof{figure}{Examples from SPHEAR, our differentiable generative 3D human head model, rendered in diverse lighting conditions using HDRI backgrounds, with diversely sampled head shapes, poses, expressions, appearance, and hair types. The resulting avatars can be easily integrated with most simulation or rendering engines and can be augmented with various other accessories or facial hair.}
\label{fig:teaser}
\vspace{8mm}
}]

\begin{abstract}
\vspace{-3mm}

We present \emph{SPHEAR}, an accurate, differentiable parametric statistical 3D human head model, enabled by a novel 3D registration method based on spherical embeddings. We shift the paradigm away from the classical Non-Rigid Registration methods, which operate under various surface priors, increasing reconstruction fidelity and minimizing required human intervention. Additionally, SPHEAR is a \emph{complete} model that allows not only to sample diverse synthetic head shapes and facial expressions, but also gaze directions, high-resolution color textures, surface normal maps, and hair cuts represented in detail, as strands. SPHEAR can be used for automatic realistic visual data generation, semantic annotation, and general reconstruction tasks. Compared to state-of-the-art approaches, our components are fast and memory efficient, and experiments support the validity of our design choices and the accuracy of registration, reconstruction and generation techniques. 
    \vspace{-5mm}
\end{abstract}

\section{Introduction}

Three-dimensional morphable models (3DMMs) of the human face and head have proven highly useful for a variety of computer vision and graphics applications. Since the introduction of the first 3DMMs \cite{blanzvetter99}, a palette of expressive models \cite{ghum2020,li2017learning,ploumpis2020towards,cao2013facewarehouse,yang2020facescape} have been developed and deployed in various contexts \cite{egger3dmms}. Nowadays, 3D face and head models serve as essential building blocks for applications such as 3D avatar creation \cite{gafni2021nerface, Khakhulin2022ROME}, media creation and editing \cite{thies2020neural, tewari2020stylerig}, 3D human reconstruction \cite{sanyal2019learning,MICA:ECCV2022}, or synthetic data generation \cite{Wood_2021_ICCV}, among others.

As the applications of 3DMMs have become more diverse and mature, the demands placed on these models have increased accordingly. However, most current models capture only the 3D geometry of the human face or head, often with limited detail, and neglect appearance entirely or provide only low-frequency texture models for the skin and face. Furthermore, eyes, teeth, the tongue, or hair are frequently missing, which are essential components in the modeling of the human head (see \cref{tab:features}). To enable the aforementioned applications, researchers palliated these missing elements by using, for example, free-form model deformation  \cite{Khakhulin2022ROME}, neural rendering \cite{gafni2021nerface}, or by relying on artist-designed assets \cite{Wood_2021_ICCV, bazavan2021hspace}.

In building an accurate 3DMM for heads, template registration is a critical first step \cite{ghum2020,li2017learning}. During registration, a predefined template is fitted to unstructured and often noisy 3D human head scans, resulting in a representation with consistent topology and semantics that can be used for subsequent statistical model learning. Therefore, since model learning is often based on registered meshes, the registration quality is crucial in achieving optimal model performance. We have found that conventional registration techniques are often insufficiently accurate and sensitive to initialization and parameterization. Therefore, manual intervention and cleaning are typically required during the registration process to develop a reliable model. These limitations are restrictive in two important ways:  \emph{First}, when parameters and initialization are not carefully chosen or manual intervention is not carefully performed, modeling accuracy suffers. \emph{Second}, the diversity and expressiveness of a model are restricted by the number of scans that can be manually inspected. To develop a model that captures the diversity of humanity, however, a large sample is preferable \cite{Booth_2016_CVPR}.

In this paper, we take 3DMMs a step further and present SPHEAR, our diverse, holistic model of the human head, that includes a color texture, surface normal maps, and a hair representation compatible with the head geometry. In addition, our model also comprises essential physical components such as eyeballs, a tongue, teeth, and matching assets for facial hair, eyebrows, and eyelashes. To enable the creation of SPHEAR, we present the following contributions: (1) A novel and robust \textbf{registration pipeline} based on a learned \textit{spherical embedding}, which provides smooth and accurate automatic registrations, agnostic to the template resolution, with preserved semantics and lower error when compared to conventional methods; 
(2) A non-linear \textbf{generative shape and expression model} for faces, learned on a newly collected 3D dataset of human facial expressions (\textbf{FHE3D}) including eye-balls, teeth and tongue; (3) A fast, flexible, neural \textbf{generative hair model}, that operates in texture space and allows for control of the number of generated strands. Moreover, our proposed Legendre polynomial encoding of hair strands enables smaller models, faster training, and a variable number of control points per strand; (4) For completeness, we also introduce a neural \textbf{generative appearance model} for detailed color and normal maps of human faces; (5) In extensive experiments, we ablate critical design choices and model variants, and demonstrate our model's accuracy, diversity, and relevance for applications.

\begin{table}
    \begin{center}
    \small
    \vspace{-5mm}
    \resizebox{0.99\linewidth}{!}{%
    \begin{tabular}{ccccccc|l}
    \rot{Non Linear} & \rot{Rigged}  & \rot{Full Head} & \rot{Eyes, Teeth, Tongue}  & \rot{Appearance} & \rot{Hair} & \rot{\# Vertices} & \\
    \hline\hline
    \xmark & \xmark & \xmark & \xmark & \cmark & \xmark & 53K & Basel FM \cite{blanzvetter99} \\

    \xmark & \xmark & \xmark & \xmark & \xmark & \xmark & 11K & Facewarehouse \cite{cao2013facewarehouse} \\

    \xmark & \cmark & \cmark & \xmark$^\dagger$ & \xmark$^\ast$ & \xmark & 5.3K & FLAME \cite{li2017learning} \\

    \xmark & \xmark & \cmark & \cmark & \xmark$^\ddagger$ & \xmark & 106K & UHM \cite{ploumpis2019combining, ploumpis2020towards} \\

    \xmark & \xmark & \cmark & \xmark & \xmark$^\ast$ & \xmark & 26.3K & Facescape \cite{yang2020facescape} \\

    \cmark & \cmark & \cmark & \xmark & \xmark & \xmark & 3.1K & GHUM \cite{ghum2020} \\
    \hline
    \cmark & \cmark & \cmark & \cmark & \cmark & \cmark & 12.1K + H & \textbf{SPHEAR (ours)} \\
    \end{tabular}
    }
    \end{center}
    \vspace{-3mm}
    \footnotesize{$^\dagger$ has eyeballs, $^\ddagger$ has a texture completion model, $^\ast$ vertex-based extension added in subsequent work}
    \caption{Overview of different features and level of detail for different face and head models. SPHEAR covers most components and is accurate at a tractable resolution. Vertices in the hair representation are not counted as their level of detail can vary, \cf \S\ref{sec:gen-hair-model}.}
    \label{tab:features}
    \vspace{-3mm}
\end{table}

\section{Related Work}

\paragraph{Registration} is the process of accurately deforming a template mesh to fit a scan, and key to learning accurate, realistic statistical head models.
Most non-rigid registration methods \cite{li2017learning, ghum2020} rely on as conformal as possible (ACAP) registration \cite{Yoshiyasu2014}, which we argue does not provide sufficient accuracy to build high quality models.
Inspired by spherical template registration \cite{gupta2020neural}, that learns a diffeomorphic flow between a sphere and a genus-0 scan, through Neural ODE integration \cite{odenet}, we go one step further and learn the registration between any genus-0 template and any genus-0 scan. 
\cite{one_ring_cheng} learn a diffeomorphism between a genus-0 sphere and the target mesh using a ResNet. They further relax the need for 3D ground truth by learning an illumination model under a reconstruction loss.  \cite{paschalidou2021neural} use an explicit family of bijective functions (i.e. \textit{normalizing flows} \cite{dinh2016density, rezende2015variational}) to model the mapping.
In this work, we show how genus-0 sphere mappings can be used for high resolution registration and triangulation, with good generalization and considerable decrease in reconstruction error. Our approach can be viewed as an intrinsic registration method, where alignment is done in the sphere embedding space. But, different from other intrinsic methods, we can easily deform between the template and the scan, in either direction. For a comprehensive survey of the non-rigid registration literature see \cite{deng2022survey}. 

\paragraph{Canonical embeddings for 3D representation} Several methods \cite{cheng2021learning, taylor2012vitruvian, groueix20183d, yang2018foldingnet, cheng2022autoregressive} have used canonical representations (e.g spheres, UV spaces, 3D codes) for embedding surfaces or point-clouds. While we share a general learning spirit with \cite{cheng2021learning}, we are the first to show how high-density face meshes can be interpolated, matched and registered at sub-millimeter reconstruction error, better than optimization-based, classical registration methods. This is achieved through \textit{explicit} spherical embeddings, with densely sampled geometry on the mesh surfaces, modulated by high-frequency Siren \cite{sitzmann2020implicit} networks. We design losses such that points can be processed independently/parallel, allowing the networks to learn the high-frequency structures.  \cite{cheng2021learning} operates solely on point-clouds ($\approx$ 2k points), is limited in the number of points it can match, and its spherical embedding formulation is mathematically different.  Instead, we operate on dense meshes for registration, where sub-millimeter accuracy is required.

\paragraph{3D morphable face and head models} aim to capture the diverse characteristics of the human face or head in form of 3D geometric models controlled by small dimensional latent parametrizations \cite{egger3dmms}. 
Initial work by Blanz and Vetter \cite{blanzvetter99} and the subsequent Basel Face Model \cite{paysan20093d, gerig2018morphable} represent only the \emph{frontal face} and ear regions and rely on PCA-based geometry and texture spaces trained on scans from 200 subjects with neutral expression.
Subsequent work enriches the shape space with more 3D scans \cite{Booth_2016_CVPR,booth20163d,booth2018large} or images from the Internet \cite{ira2013wild3dmm, Booth_2017_CVPR, DECA:Feng:SIGGRAPH:2021}.
Disentangled facial identity and expressions are obtained by bilinear models \cite{vlasic2006face, cao2013facewarehouse, BolkartWuhrer2015_groupwise, yang2020facescape}. 
More related are \emph{full head} models, typically trained on a larger and more diverse set of subjects \cite{li2017learning, ghum2020, yang2020facescape, ploumpis2019combining, ploumpis2020towards}, and allowing for separate control over geometry, pose, and expression.
While most use 3D scans and sometimes images, SCULPTOR \cite{qiu2022sculptor} further utilizes CT scans.
Deep neural networks have been used to obtain richer non-linear latent representations \cite{bagautdinov2018modeling, abrevaya2018multilinear, ranjan2018generating, tran2018on, tran2018nonlinear, tran2019towards, ghum2020}.
Finer textures and sometimes normal maps are obtained using GANs \cite{goodfellow2014} -- however, mostly for texture reconstruction or completion \cite{Luo_2021_CVPR,ploumpis2020towards,dynamicfacialasset2020} and seldom for statistical appearance modeling \cite{li2020learning}.
Neural fields \cite{xie2022neural} have been explored more recently \cite{zheng2022imface, yenamandra2021i3dmm, giebenhain2022nphm, zanfir2022phomoh}, sometimes focusing on image quality \cite{hong2022headnerf, wang2022morf} rather than generative 3D modeling.

\begin{figure}
\begin{center}
    \includegraphics[width=0.8\linewidth]{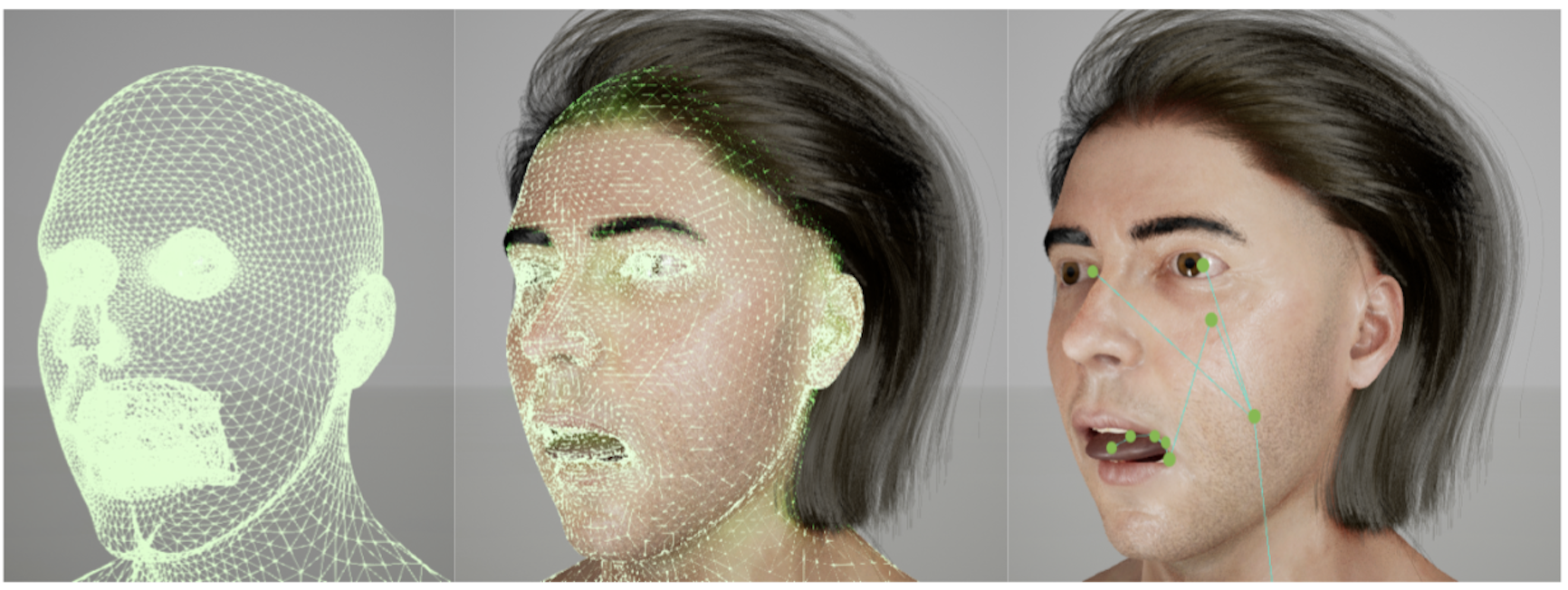}
\end{center}
\vspace{-4mm}
\caption{Template model used to create SPHEAR. From left to right, the base geometry including the tongue, eyeballs, and teeth, the hair, the template mesh connectivity, and the kinematic rig (joints).}
\label{fig:template_model}
\end{figure}

\begin{figure}[t]
\center{\includegraphics[width=1.\linewidth]{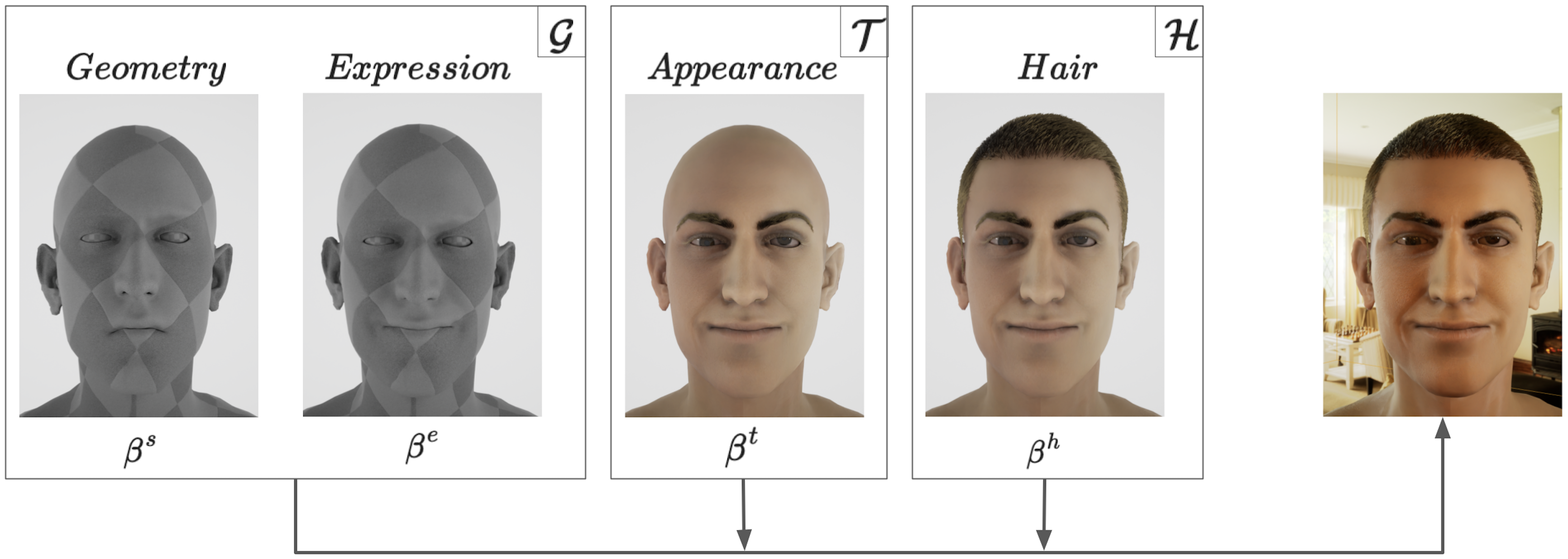}}
\caption{Overview of our holistic head model generation pipeline. Latent $\bbeta^{s}$ and $\bbeta^{e}$ of shape and expression encodings are passed through 3D generative VAE models, to obtain high-resolution meshes with a consistent topology. We sample $\bbeta^{t}$ and pass it through our generative texture model $\mathcal{T}$ in order to add detailed appearance and enable the creation of PBR materials. We sample $\bbeta^{t}$ from the latent space of our generative model $\mathcal{H}$ to add detailed strand-based hair. We obtain an animatable avatar which is compatible with most modern rendering engines and can be placed in scenes under various lighting conditions. See \cref{fig:teaser} and \cref{fig:variations} for illustration.}
\end{figure}\label{fig:overview_figure}

\paragraph{Personalized avatar creation} is the process of computing photorealistic models of individuals given a single image or a video \cite{park2021nerfies, Mihajlovic:KeypointNeRF:ECCV2022,burkov2022multineus,ramon2021h3d}, sometimes even for the full body \cite{Saito_2019_ICCV, alldieck2022phorhum, corona2022s3f, Huang_2020_CVPR}. 3DMMs are sometimes the base representation  \cite{Khakhulin2022ROME,zheng2022imavatar, athar2022rignerf}, which allows for animation. However, the generative identity space is lost in this process. Our SPHEAR could be used in a personalization process, either as a statistical prior or as a base representation to further refine, but that is not the goal of this paper.

\paragraph{Hair modeling} is critical for many generative tasks, however current 3DMMs lack it. Standalone hair models based on generative deep neural networks have been introduced \cite{hairnet2018Eccv, Olszewski_2020_CVPR, Saito:2018:HSU:3272127.3275019, zhang2019hair} without an immediate connection to a generative head model. Laborious work is required to combine the two \cite{Wood_2021_ICCV}.
Other methods leverage neural rendering \cite{rosu2022neuralstrands} and require a specialized capture system with hundreds of cameras for accurate reconstruction and rendering. We aim to statistically generate a variety of hairstyles given training data, and to be able fit to various input signals, within flexible memory and time budgets.

\section{Methodology}

We introduce SPHEAR, a 3D articulated head model which contains generative shape, expression, hair, and appearance components, making it the first complete statistical head model to date. We start from an artist-defined rigged template mesh, $\mathbf{T} = \{\mathbf{V}_t, \mathbf{F}_t\}$ where $\mathbf{V}_t \in \mathbb{R}^{12201 \times 3}$ is the set of vertices and $\mathbf{F}_t \in \mathbb{N}^{24318 \times 3}$ is the set of faces. The template has $J=15$ joints which also control the tongue, left and right eyeballs, and lower and upper eyelids. The initial set of skinning weights is $\bomega_{0} \in \mathbb{R}^{12201 \times 15}$.
Formally, SPHEAR is controlled by a set of low-dimensional embedding vectors and pose parameters
\small
\begin{equation*}
\mathcal{M}\left(\bbeta^{s}, \bbeta^{e}, \btheta, \bbeta^{t}, \bbeta^{h} \right) = \left\{ \mathcal{G}(\bbeta^{s}, \bbeta^{e}, \btheta ), \mathcal{T}(\bbeta^{t} ), \mathcal{H}( \bbeta^{h}  )\right\}.
\end{equation*}
\normalsize
The geometric component $\mathcal{G}$ generates meshes with the same number of vertices and connectivity as the template. It is controlled using the shape encoding $\bbeta^{s}$ for the neutral expression,  $\bbeta^{e}$ encodes facial expressions in neutral head pose, and $\btheta$ is a vector containing the pose parameters and can be represented as Euler angles, rotation matrices, or 6D representations \cite{zhou2018continuity}. The appearance component $\mathcal{T}$ generates high resolution color and normal textures from a latent representation $\bbeta^{t}$. The hair component $\mathcal{H}$ generates dense strands from a low-dimensional $\bbeta^{h}$ code.   

To learn different model components, we collect a new 3D dataset of facial human expressions, \textbf{FHE3D}, containing 240K photorealistic scans captured from more than 600 diverse subjects with a proprietary system operating at 60Hz. For each subject, we capture 40 different facial expression tracks lasting 2 seconds each.
For each scan, our system provides 3D geometry as well as texture maps. We keep a held out test set of 1,000 scans from 25 subjects, each with 40 different facial expressions (\textbf{FHE3D-TEST}).

In the sequel, we will describe the creation of SPHEAR, starting with the process of template registration to captured scans. We will then detail how we create the various components $\mathcal{G}, \mathcal{T}, \mathcal{H}$.

\subsection{Sphere Embeddings for Accurate Registration}
\label{sec:sphere_emb}

We call \textit{spherical embedding} a diffeormorphism (\ie invertible and smooth mapping) between the unit 2-sphere $\mathbb{S}^2$ and any smooth 2-manifold of genus-0 $\mathbb{M}$, $g:\mathbb{S}^2 \rightarrow \mathbb{M}$. The sphere is a strong regularizer to represent genus-0 topologies, and one can easily and efficiently parameterize them. For our use case, head meshes can be approximated as genus-0 topologies. We propose to reparameterize each training head scan as a sphere, by learning the mapping $g$ and its corresponding approximate inverse function $f:\mathbb{M}\rightarrow \mathbb{S}^2$. For all 3D points $\mathbf{x} \in \mathbb{M}$, then the functions have to satisfy $\mathbf{x} \approx f(g(\mathbf{x}))$. To regularize the solution, we discourage sharp changes in gradients by adding a penalty on the second-order derivatives of $g$. This reparameterization can be used for shape learning, interpolation and for establishing semantically meaningful correspondences. We will apply these properties in the context of registration.

\begin{figure}[t]
    \centering
    \includegraphics[width=.8\linewidth]{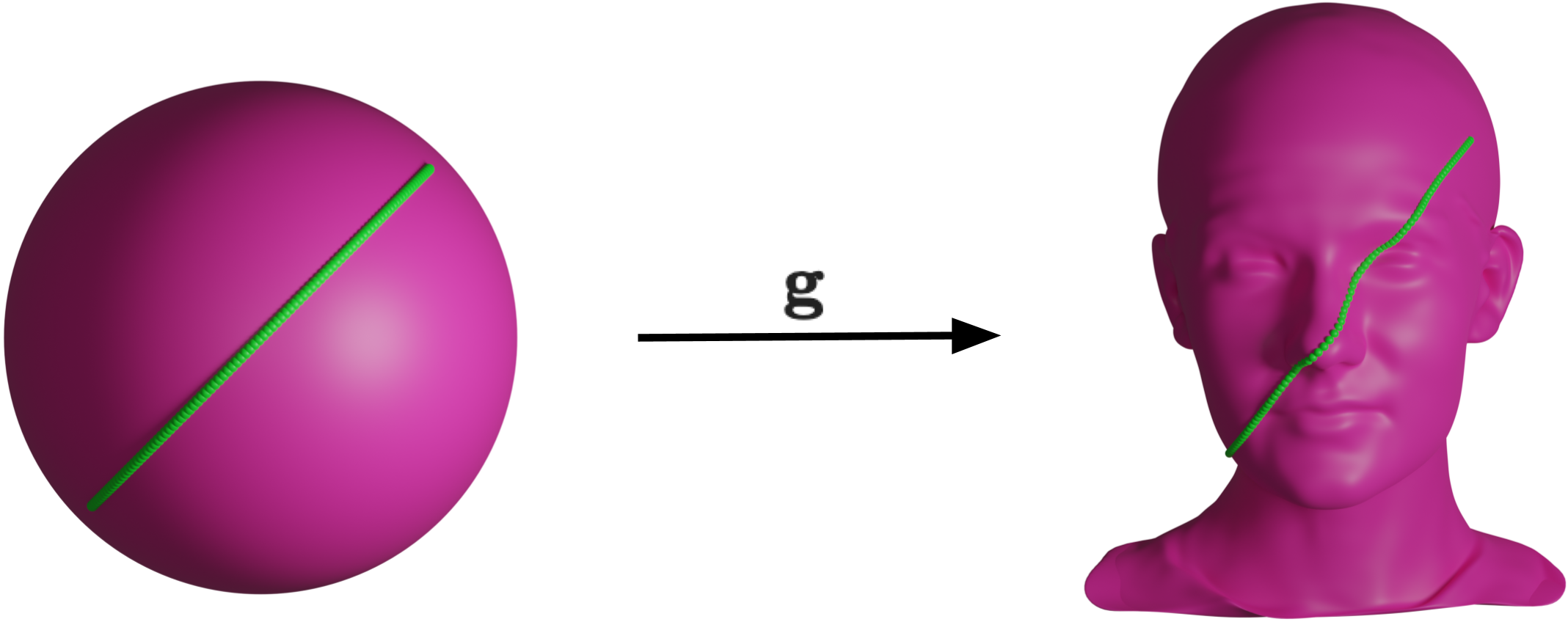}
    \caption{\textbf{Manifold equivalency} We learn a spherical mapping of a scan, and walk on the surface of the sphere (left). Through the learned function $g$, we show how we can walk on the surface of the scan (right). The figure on the right is not the scan itself, but the decoded $g(\mathbf{Y})$, where $\mathbf{Y}$ are points on an icosphere of depth 7.}
    \label{fig:sphere_walking}
    
\end{figure}

In contrast to our formulation, previous approaches enforce the invertibility property \cite{gupta2020neural, one_ring_cheng, paschalidou2021neural}, either by using a family of bijective functions \cite{paschalidou2021neural} or by integration over a smooth vector field \cite{gupta2020neural, one_ring_cheng}. The invertibility property guarantees perfect reconstruction loss (because $g: \mathbb{R}^3 \rightarrow \mathbb{R}^3$ is bijective and $f=g^{-1}$), but does not guarantee that the function is a bijection \textit{between the desired manifolds}. The training loss is thus limited to either Chamfer distances or other indirect reconstruction losses, in order to force both manifolds to match the expected $\mathbb{S}^2$ and $\mathbb{M}$. We argue that these approaches cannot capture the high-frequency details of head scans and limits potential applications where we would like to condition independently $g$ and $f$. Please see Fig. \ref{fig:sphere_walking} for an example.

\paragraph{Spherical registration.} Assuming we have a target scan $\mathbf{S} = \{\mathbf{V}_s, \mathbf{F}_s\}$, with arbitrary vertices and triangulation, we want to \textit{register} the template $\mathbf{T}$ to $\mathbf{S}$. This means explaining the geometry of the target by a deformation of the template in a semantically meaningful sense. This is an ill-posed problem, with many regularizations and procedures explored in the literature (ARAP \cite{sorkine2007rigid}, ACAP \cite{Yoshiyasu2014}, \etc).

Our solution is to embed both the target and the template in the \textbf{same} spherical embedding space, where they must be aligned. To do so, we augment two Siren \cite{sitzmann2020implicit} activated MLPs, $S_e$ and $S_d$ with two additional modulator networks \cite{mehta2021modulated}, $S_e^m( \mathbf{c}, \bphi^m_e), S_d^m(\mathbf{c}, \bphi_d^m)$. Here, $\mathbf{c}$ is an input code to the modulation networks, while $\bphi^m_e, \bphi^m_d$ are trainable parameters. The codes will be used to modulate the Siren-based encoders and decoders, \ie $S_e(*, \bphi_e) \odot S_e^m( \mathbf{c}, \bphi^m_e)$ and $S_d(*, \bphi_d) \odot S_d^m(\mathbf{c}, \bphi_d^m)$.  The encoder $f$ and decoder $g$ will have the following forms 
\begin{align}
    f(\xb, \bphi_e, \bphi^m_{e}, \mathbf{c}) &= \mathbf{\Pi}\left(\xb + S_e(\xb, \bphi_e) \odot S_e^m( \mathbf{c}, \bphi^m_e)\right) \\
    g(\yb, \bphi_d, \bphi^m_{d}, \mathbf{c}) &= \yb + S_d(\yb, \bphi_d) \odot S_d^m(\mathbf{c}, \bphi_d^m)
\end{align}
where $\mathbf{\Pi}$ is a normalization operator projecting all entries onto the unit 3D sphere. We remove explicit dependencies on $\bphi_{e}, \bphi^m_{e}, \bphi_{d}, \bphi^m_{d}$ for clarity.

For our task, we use $2$ randomly generated codes, $\mathbf{c}_t$ and $\mathbf{c}_s$, for both the template and the target scan. During training, we will embed the template $\mathbf{T}$ and the scan $\mathbf{S}$ into the spherical space, through $f(*, \mathbf{c}_t)$ and $f(*, \mathbf{c}_s)$, and decode them using $g(*, \mathbf{c}_t)$ and $g(*, \mathbf{c}_s)$, respectively. To obtain aligned surfaces $\mathbf{T} \rightarrow \mathbf{S}$, at inference time, we will decode $g(f(\mathbf{T}, \mathbf{c}_t), \mathbf{c}_s)$. For this to work, both mesh surfaces must be aligned in the \textbf{spherical space}. To enforce alignment, we use landmarks attached to each of the two meshes. We define $\Pb^s \in \mathbb{R}^{478\times 3}$ for the scan landmarks and $\Pb^t \in \mathbb{R}^{478\times 3}$ for the template, and guide them to be consistent in the spherical embedding spaces. In order to obtain $\Pb^s$ we render $\mathbf{S}$ in 16 synthetic views, with known camera parameters, and use FaceMesh \cite{grishchenko2020attention} to obtain 478 2D landmarks $\Pb_{2d} \in \mathbb{R}^{16 \times  478 \times 2}$. We triangulate these points (through nonlinear optimization) to obtain the desired 3D landmarks. For $\Pb^t$ we use predefined semantic locations on the template $\mathbf{T}$.

To train our networks, we sample $N_{s}$ points on the scan surface -- \ie $\Xb^s \in \mathbb{R}^{N_s\times3}$ -- and $N_{s}$ points on the surface of the template -- \ie $\Xb^{t} \in \mathbb{R}^{N_s\times3}$ -- and reconstruct them using our encoder/decoder networks. The landmarks are embedded into the sphere and matched together. This amounts to the following losses
\begin{align*}
    \widetilde{\mathbf{\Xb}}^t_i = g(f(\Xb^t_i, \mathbf{c}_t), \mathbf{c}_t),
    \quad \widetilde{\mathbf{\Xb}}^s_i = g(f(\Xb^s_i, \mathbf{c}_s), \mathbf{c}_s)
\end{align*}
\begin{align*}
    \mathbf{\Qb}^t_i &= f(\Pb^t_i, \mathbf{c}_t),
    \quad \mathbf{\Qb}^s_i = f(\Pb^s_i, \mathbf{c}_s) \\
    \mathcal{L}_{rec} &=  \sum_{i} \|\Xb^t_i - \widetilde{\mathbf{\Xb}}^t_i\|_2 + \|\Xb^s_i - \widetilde{\mathbf{\Xb}}^s_i\|_2 \\
    \mathcal{L}_{lmks} &= \sum_{i} \|\mathbf{\Qb}^t_i - \mathbf{\Qb}^s_i\|_2. 
\end{align*}
To ensure a smooth solution, we introduce a second-order regularization term, defined as the squared Frobenius norm of the Hessian matrix of $g$, $\ie$ $\Hb_g$. We sample $N_e$ points $\Yb \in \mathbb{R}^{N_e\times3}$ on the surface of the sphere and then compute the regularization term $\mathcal{L}_{reg} = \sum \|\Hb_g(\Yb)\|_F^2.$

\begin{figure*}
\begin{center}
    \includegraphics[width=0.99\linewidth]{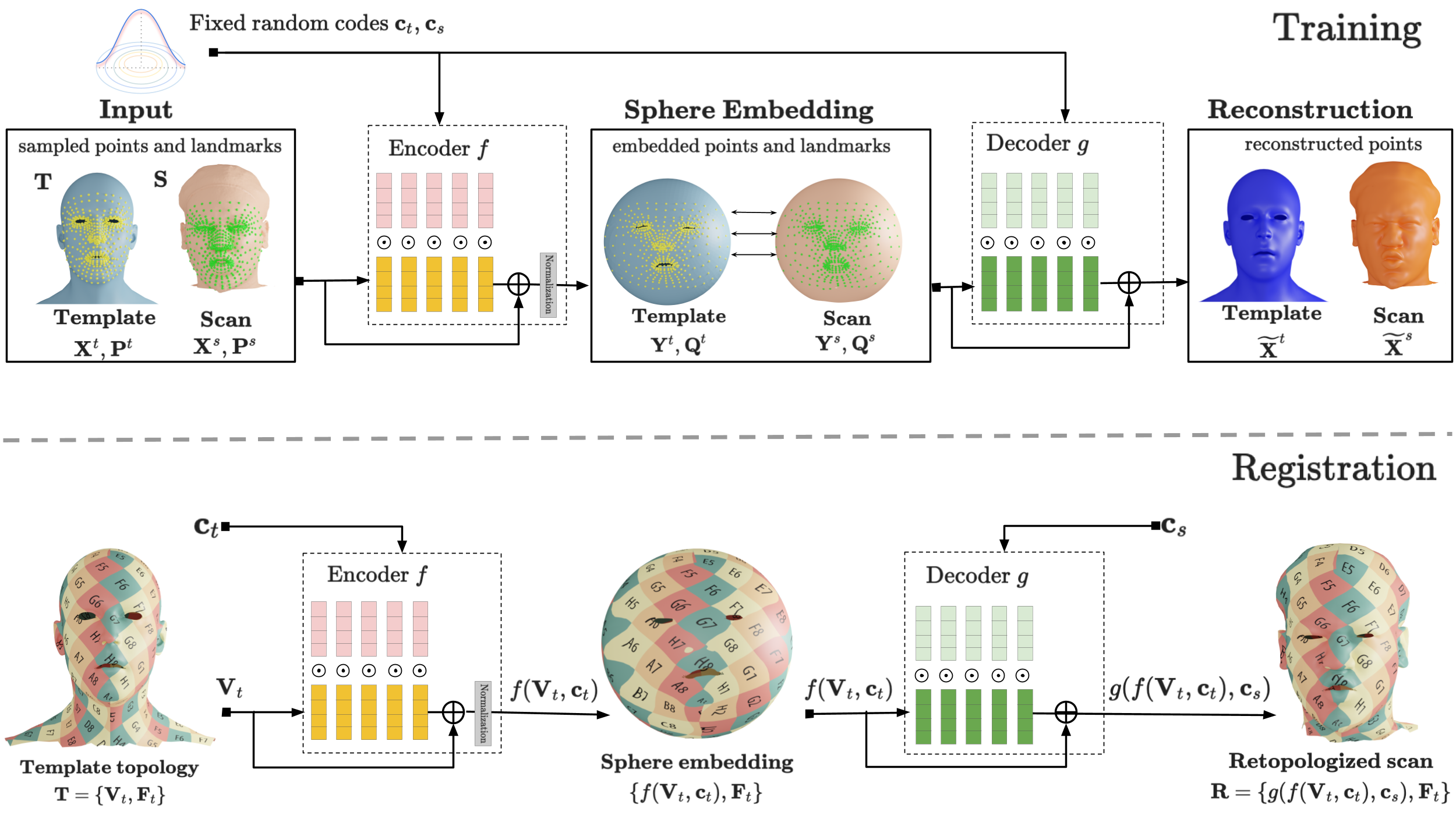}
\end{center}
\vspace{-4mm}
\caption{Overview of our registration pipeline. \textbf{Training.} During training, we learn a joint spherical embedding of a template mesh $\mathbf{T}$ and a target scan $\mathbf{S}$, with  3D points $\textbf{X}^{t}$ and $\textbf{X}^{s}$, sampled on the surface, respectively. The two meshes have associated input landmark locations $\mathbf{P}^{t}$ and $\mathbf{P}^{s}$, with common semantics. Fixed random codes $\cb_s$ and $\cb_t$ inform the encoder $f$ and decoder $g$ networks about the identity of points. The network maps to and from the spherical space, with reconstructions $\widetilde{\Xb} = g(f(\Xb, \cb), \cb)$ that should match the input points $\Xb$. To regularize the reconstruction and ensure smoothness, we add a 2nd-order penalty for the generator $g$ as the trace of its Hessian matrix. To align the template and target, we enforce landmark consistency in the spherical embedding space, \ie $\mathbf{Q}^t = f(\mathbf{P}^t, \mathbf{c}_t)$ should be equal to $\mathbf{Q}^s = f(\mathbf{P}^s, \mathbf{c}_s)$. Note that $\textbf{X}^t$ and $\textbf{X}^s$ are sampled from their corresponding meshes at each training step, ensuring fine-grained, detailed reconstruction. \textbf{Registration.} Once the networks have been trained, we can now swap codes for the generator: we embed the template topology on the sphere via $f(*, \cb_t)$, but decode via $g(*, \cb_s)$. Using a UV checker texture, we show that the semantics are preserved when going from template to target. }
\label{fig:spherical_registraion_diagram}
\end{figure*}

We add this regularization loss for both surfaces. The total loss becomes $\mathcal{L}_{sr} = \mathcal{L}_{rec} + w_{lmks} \mathcal{L}_{lmks} + w_{reg} \mathcal{L}_{reg}$. Once trained, we can obtain the topology of either template or scan, but on the surface of the other. For example, to retrieve the template registered to the target scan, we simply compute
\begin{align}
    \mathbf{R} = \left\{\mathbf{V}_{r}:g(f(\mathbf{V}_t, \mathbf{c}_t), \mathbf{c}_s), \mathbf{F}_t)\right\}.
\end{align}
See \cref{fig:spherical_registraion_diagram} for a visual illustration of the method.

\begin{figure}
\begin{center}
    \includegraphics[width=0.8\linewidth]{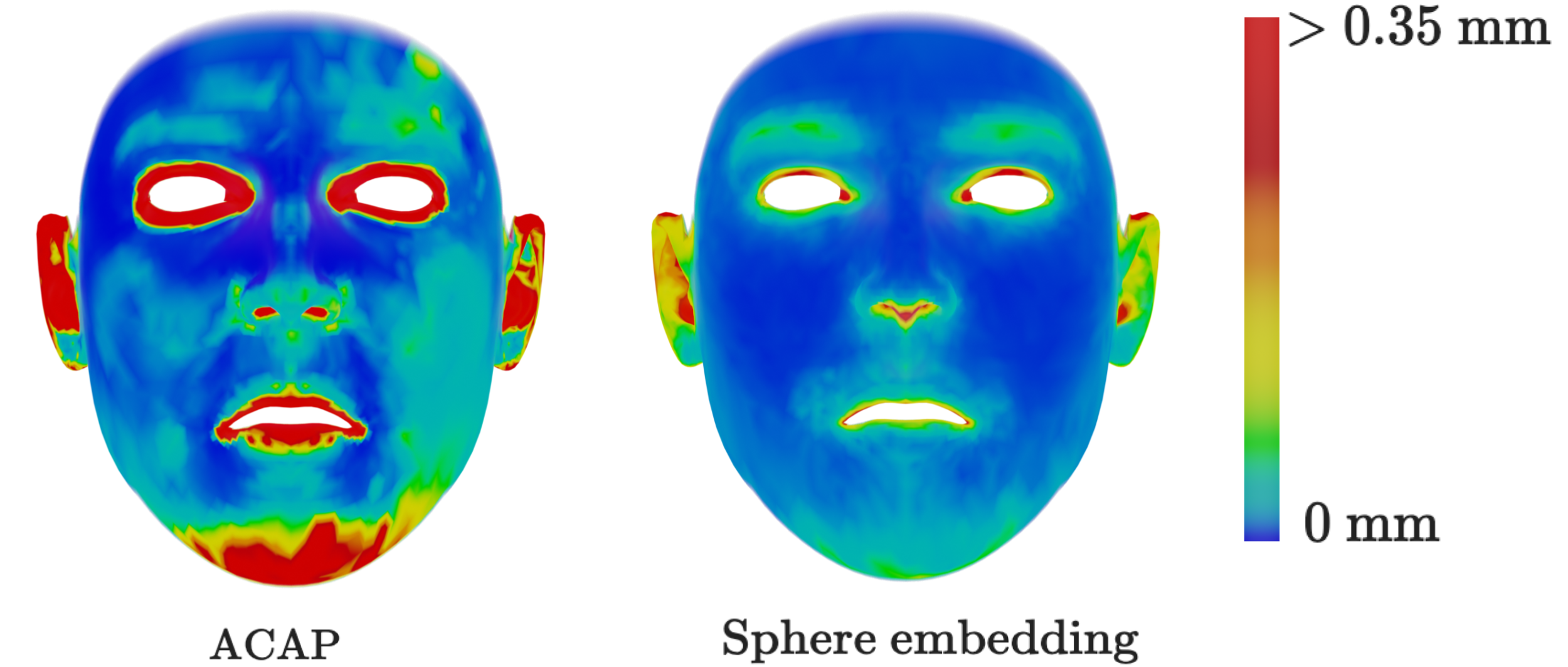}
\end{center}
\vspace{-2mm}
\caption{Quantitative evaluation of our spherical registration pipeline. We compare the template-to-scan ICP error with ACAP \cite{Yoshiyasu2014} on the \textbf{FHE3D} dataset. On average, our method achieves $0.17$ mm per-vertex error compared to $0.44$ mm for ACAP.}
\label{fig:reg_qualitative}
\vspace{-2mm}
\end{figure}

We train and test the registration pipeline, on the face region (including ears), against ACAP. We show averaged template-to-scan ICP error maps, across all training examples, in \cref{fig:reg_qualitative}. The mean errors are $0.17$ mm for our method and $0.44$ mm for the classical ACAP registration. See the additional experiments in \S\ref{sec:experiments}.

\begin{figure}
\begin{center}
    \includegraphics[width=0.8\linewidth]{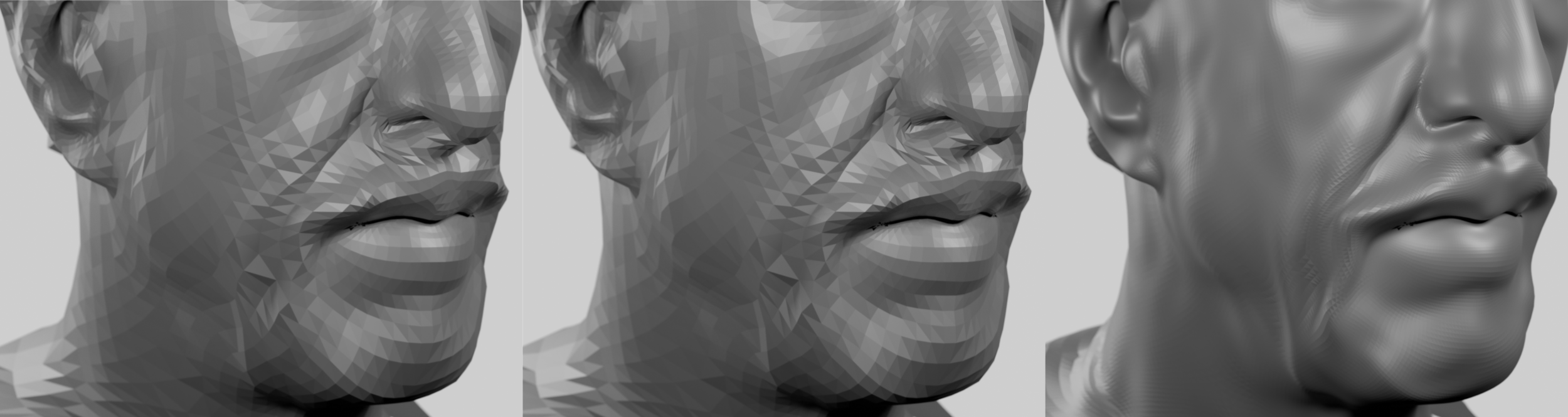}
\end{center}
\vspace{-4mm}
\caption{\textbf{Tessellation invariant sphere embedding}. \textbf{Left.} The template topology passed through the registration pipeline, \textit{after training}. \textbf{Center.} Simulated re-tessellation of the template ($\approx 16\times$ more vertices and faces), by face sub-division of registered result from \textit{Left}. Note how this does not add any new details.
\textbf{Right.} Re-tessellated template passed through the registration pipeline, \textit{after training}. Finer details are added because of the shared spherical embedding with the surface of the target scan.}
\label{fig:spherical_subdivision}
\end{figure}

\noindent\textbf{Tessellation invariant registration.} Because our training pipeline maps points \textbf{sampled on the surfaces}, we are invariant to any particular tessellation of the input meshes. Our approach allows the registration of more detailed meshes to the target scan as long as these are aligned to the same level set as the original input template. Thus we can get new registrations of higher resolution meshes with only one feedforward pass through our spherical embedding encoders and decoders. For illustration, see \cref{fig:spherical_subdivision}, where we show how a $16\times$ subdivided template $\mathbf{T}^\prime = \{\mathbf{V}^\prime_t, \mathbf{F}^\prime_t\}$ is used instead. Note that we do not require any new retraining, which makes the registration pipeline flexible to any new tessellation, vertex or face re-orderings.

\begin{figure}[ht!]
    \vspace{-2mm}
    \centering
    \includegraphics[width=1.\linewidth]{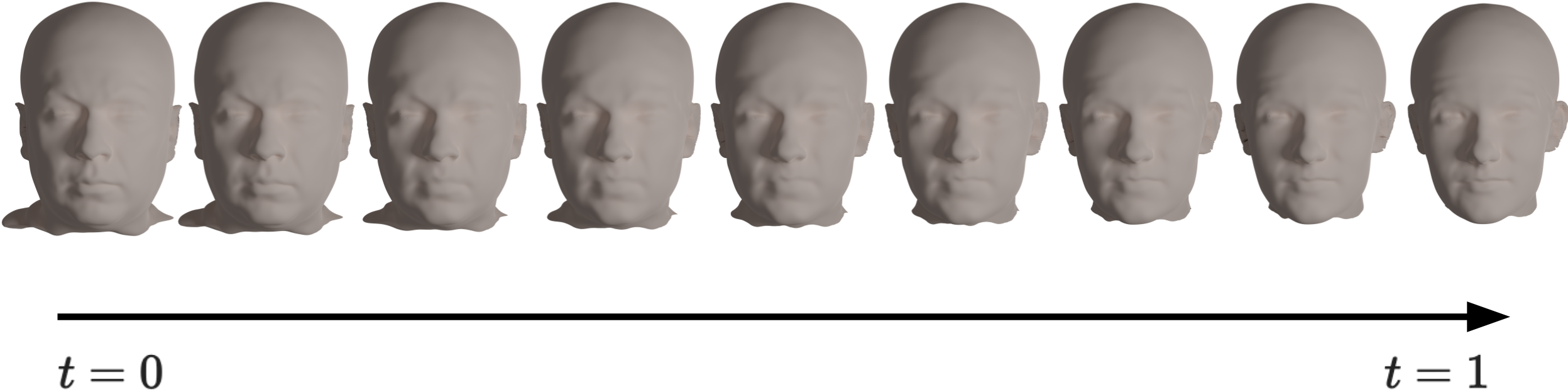}
    \caption{From left to right: interpolation between two latent codes associated with two scans, and decoded through $g$. Notice the smooth transition and the maintaining of semantics (e.g. nose shrinks accordingly).}
    \label{fig:sphere_interpolation}
    \vspace{-4mm}
\end{figure}

\paragraph{Code semantics.} We do another experiment in which, instead of a template $\mathbf{T}$ and a scan $\mathbf{S}$, we use two scans $\mathbf{S}_1$ and $\mathbf{S}_2$, with associated latent codes $\cb_{s1}$ and $\cb_{s2}$. To investigate the semantic meaning of these codes, we replicate the registration design for these new inputs. We then embed the vertices of $\mathbf{S}_1 = \{\Vb_s^1, \Fb_s^1\}$ into the sphere by $\Yb_{s1} = f(\Vb_s^1, \cb_{s1})$, but decode by $g(\Yb_{s1}, (1 - t)\cb_{s1} + t \cb_{s2})$, where $t$ is an interpolation timestep between $0$ and $1$. We show the interpolation result in \cref{fig:sphere_interpolation}. Notice how the reconstruction is a smooth transition from one scan to the other, while maintaining semantical knowledge about different face parts.

\paragraph{Secondary components.} Secondary components, such as eyes and teeth, are only partially captured by the scanner. The eyeballs are initialized using the triangulated positions of the iris landmarks \cite{grishchenko2020attention}. Their position is further optimized so that we minimize eyeball intersections with the rest of the scan for all expressions of a given subject. We perform a similar operation for the position of the teeth. More detail is given in the Sup. Mat.

\subsection{Generative Shape and Expression Model}

\begin{figure}
\begin{center}
    \includegraphics[width=.9\linewidth]{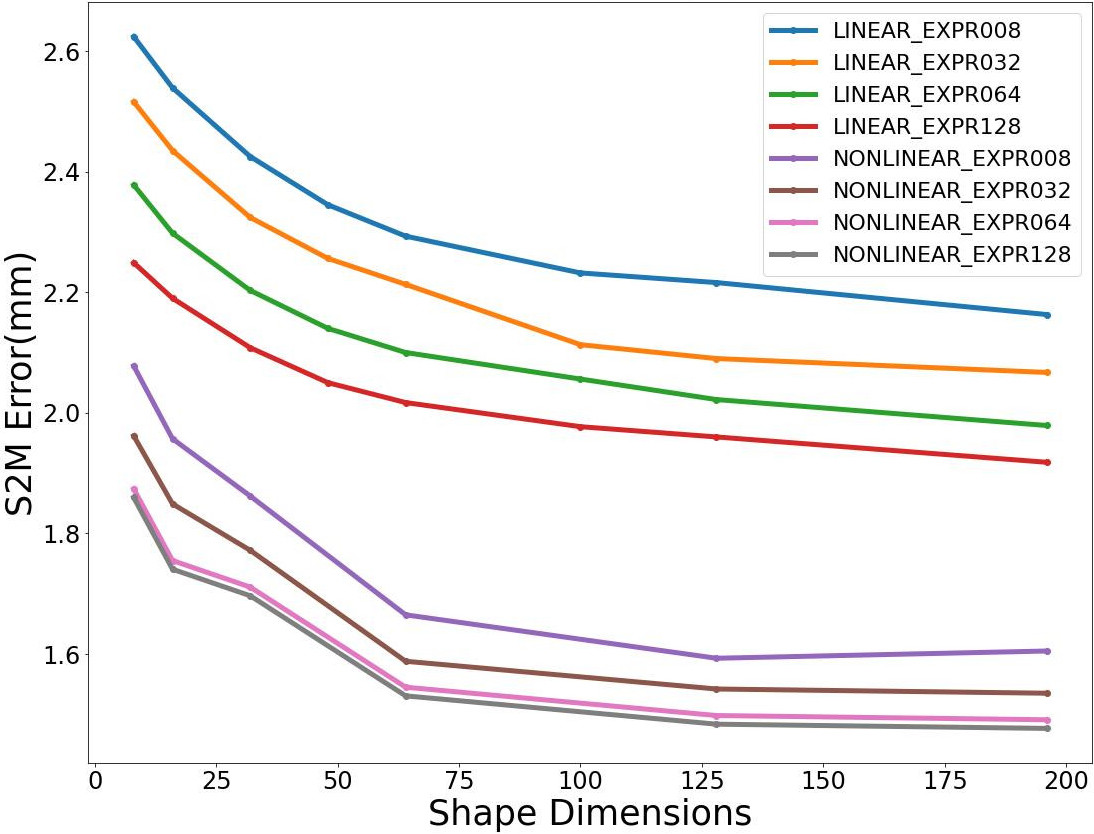}
\end{center}
\vspace{-4mm}
\caption{We ablate the SPHEAR model to compare linear versus non-linear latent representations. We evaluate on \textbf{FHE3D-TEST} set and report scan to mesh errors after fitting different models to scans. We observe a clear advantage for non-linear representations, which is consistent with previous findings \cite{ghum2020}. }
\label{fig:ablations_linear_versus_nonlinear}
\end{figure}

We employ the proposed spherical registration pipeline to register the template mesh on the \textbf{FHE3D} dataset. Additionally, we enhance the shape diversity by incorporating data from the \textbf{CAESAR} \cite{robinette2002civilian} dataset, which comprises 4,329 subjects annotated with facial landmarks. Following the approach in \cite{ghum2020}, we train an end-to-end head model that includes a shape embedding $\bbeta^{s}$, a facial expression embedding $\bbeta^{e}$, skinning weights $\bomega$, a shape-dependent joint centers regressor, and pose-space deformations, which depend on the head pose $\btheta$. The geometric model $\mathcal{G}\left(\bbeta^{s}, \bbeta^{e}\right)$ and its main sub-components are illustrated in Fig.~\ref{fig:overview_figure}. Both the shape and expression embeddings are modeled using a variational auto-encoder. The former is trained on disarticulated neutral shapes, while the latter is trained on disarticulated shapes with different expressions. We conduct an ablation study in ~\cref{fig:ablations_linear_versus_nonlinear}, where we analyze the effects of varying the number of latent dimensions for the two embeddings, and additionally compare against a linear version based on PCA. We report the scan to mesh (S2M) fitting errors on the \textbf{FHE3D-TEST} set. We observe that errors saturate around $64$ dimensions and that further increasing embedding size brings little to no improvement. As a result, we also use $64$ dimensions for our embeddings in practice. Furthermore, the non-linear models show significant lower errors than the linear counterparts, validating our approach. 


\subsection{Generative Hair Model}\label{sec:gen-hair-model}

To train our hair generation module $\mathcal{H}$, we use a collection of roughly $350$ hairstyles \cite{hu2015single} with an average of $10,000$ hair strands, each with $100$ 3D coordinates of the control points. Our goal is to build a latent, generative space from which we can draw hair samples that are diverse, smooth, and intersection-free with the head mesh. We aim for generation speed, memory and back-propagation efficiency for fitting to different input modalities.

For each training sample, we represent its hair strands as a texture map $\mathbf{U}_{h} \in \mathbb{R}^{256 \times 256 \times 100 \times 3}$ for the scalp, where $100$ is the number of control points starting from the root, and $256 \times 256$ is the resolution of the scalp texture. In practice, each hair strand is encoded in its tangent-bitangent-normal representation with respect to the starting triangle on the scalp. To improve memory requirements,  we approximate the hair strands with parametric curves modeled as degree $5$ Legendre polynomials with \textit{ortho-normal} bases. We obtain a new representation $\mathbf{U}_{hl} \in \mathbb{R}^{256 \times 256 \times 3 \times 6}$, where $6$ is the number of coefficients per polynomial. \textbf{To build a generative space}, we use variational auto-decoders \cite{rezende2015variational, park2019deepsdf} where each training sample will have a trainable mean $\bbeta^{h} \in \mathbb{R}^{16}$ and log-variance $\sigma \in \mathbb{R}^{16}$, where $16$ is the latent embedding size. Using the re-parameterization trick \cite{kingma2013auto} we get a code $\zz = \bbeta^{h} + \eta \exp\left({\sigma}\right)$, where $\eta$ is sampled from $\mathcal{N}\left(\mathbf{0}, \mathbf{I} \right)$. We use $KL$-divergence during training as a regularizer. For the network architecture, we use a similar approach to \cite{karras2017progressive}, where every code embedding $\zz$ will create an initial map, gradually de-convoluted to match the target hair texture. See the Sup. Mat. for more details on the network architecture and representation.

\begin{figure}
\begin{center}
    \includegraphics[width=.8\linewidth]{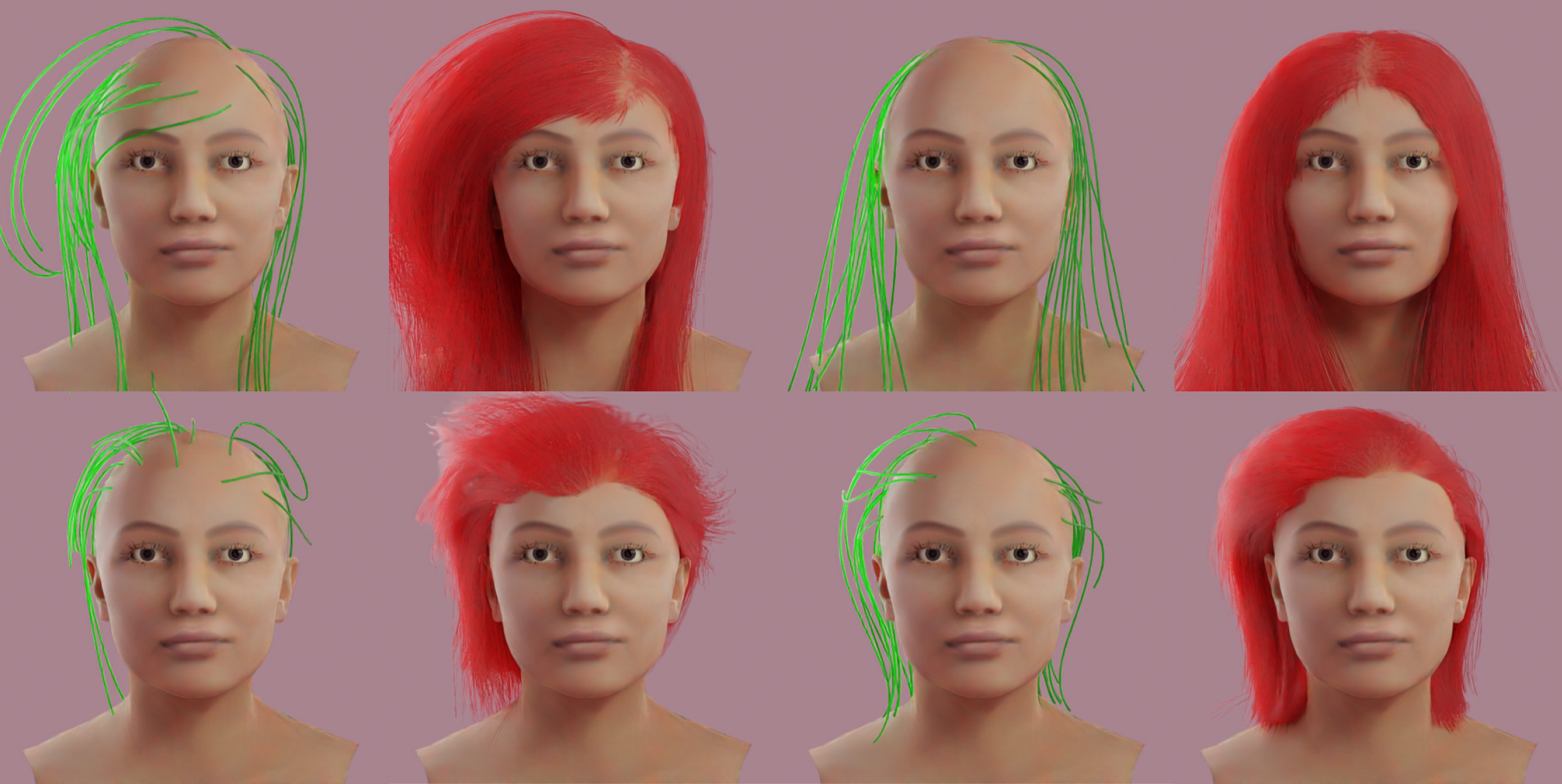}
\end{center}
\vspace{-4mm}
\caption{Hair fitting to strands guides. Green: the input strands (maximum 50). Red: reconstructed hair geometry. Optimization minimizes the distance between model predictions $ \mathcal{H}\left(\bbeta^{h} \right)$ and target hair strands, with regularization on $\bbeta^{h}$.}
\label{fig:hair_fitting_to_strands}
\end{figure}

\begin{figure}
\begin{center}
    \includegraphics[width=\linewidth]{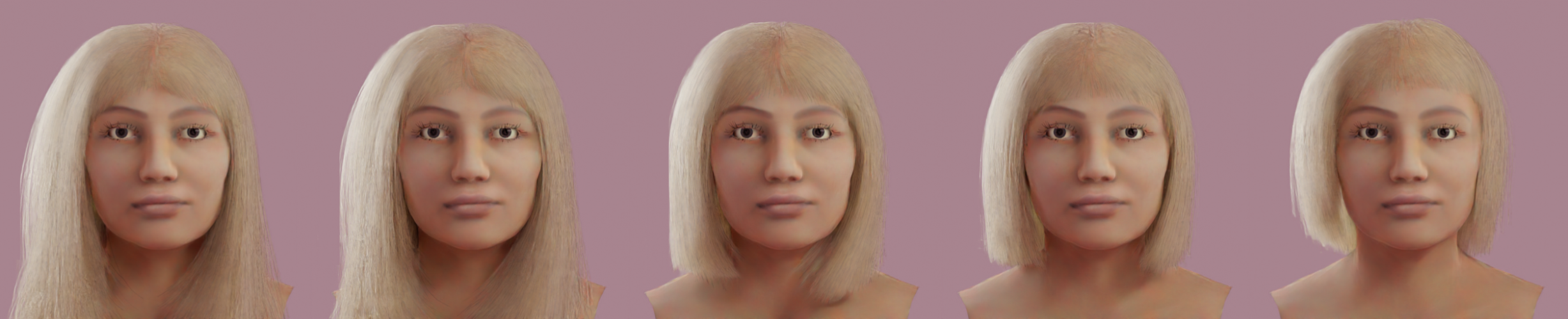}
\end{center}
\vspace{-4mm}
\caption{From left to right: interpolation in latent space for the hair generator $\mathcal{H}$, between two (start and end) randomly generated codes. Notice smooth results and plausible hair reconstructions. Here we up-sampled the scalp textures $2\times$ and evaluated $20$ control points for each Legendre polynomial, for an average of $24,000$ strands and $480,000$ control points.}
\label{fig:hair_interpolation_diagram}
\vspace{-2mm}
\end{figure}

\noindent \paragraph{Properties} After training, we generate samples running $ \mathcal{H}\left(\bbeta^{h} \right)$, where $\bbeta^{h}$ is sampled from a standard normal distribution. In \cref{fig:hair_interpolation_diagram}, we show that the learned latent space has semantics, by interpolating between $2$ randomly generated codes and showing the reconstructed hair. Because our method generates hair in a texture space, we can resize the maps to control the desired number of strands. Because the strands are encoded using Legendre polynomial bases, we can evaluate for any number of control points. This gives us flexibility in generating plausible hair styles for various computational budgets.
Different from \cite{rosu2022neuralstrands}, which learns a latent space for each strand \textit{independently}, we train a generative model for all our strands at once, basically being able to generate consistent hairstyles, not hair-strands. We also do not use Siren networks or modulators, or VAEs. Compared to variational volume encoding networks such as \cite{Saito_2019_ICCV, zhang2019hair}, which are more involved and more accurate, we are much faster during inference: those methods decode in roughly $1$ second, while we do it in roughly $1$-$2$ milliseconds. To show the practical application of our decoder architecture, we ran fitting experiments to various hair guides using BFGS. An experiment finished in under $5$ seconds, which amounts to $~1.5k$ network forward + backward passes. Results can be seen in Fig. \ref{fig:hair_fitting_to_strands}.


\subsection{Generative Appearance Model}

\begin{figure}
\begin{center}
    \includegraphics[width=0.90\linewidth]{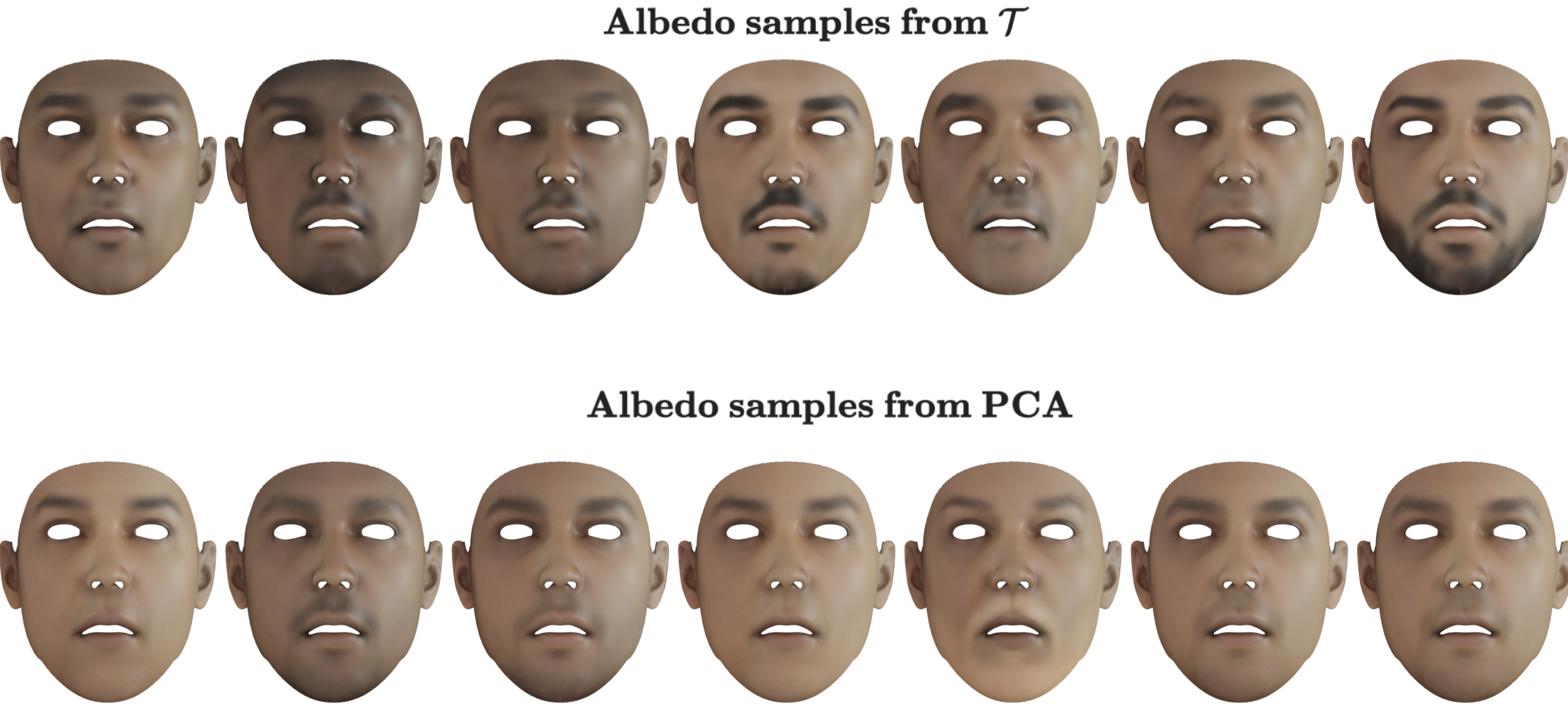}
\end{center}
\vspace{-3mm}
\caption{Qualitative comparison between PCA and our generative texture model $\mathcal{T}$. For the same code embedding size, we generate random samples and show them in two rows. Note the increased variability in appearance generated by our method (\eg sharper beards, darker skin tones, diverse eyebrow shapes, \etc), that uses $\approx50$ times less parameters than PCA.}
\label{fig:pca_vs_generative_qualitative_comparison}
\vspace{-1mm}
\end{figure}

For completeness, we also train a generative face appearance model, similar in architecture to the hair network. To make it lighter, we downsample the target maps to $1024 \times 1024$ resolution. At test time, we sample $\bbeta^{t} \in \mathbb{R}^{32}$, where $32$ is the latent embedding size, from the standard normal distribution and obtain $\mathcal{T}\left( \bbeta^{t} \right)$. We post-process by up-sampling using the super-resolution algorithm of \cite{wang2018esrgan} in order to obtain $4k$ maps. We supervise training using an Euclidean distance loss between the ground truth and predicted maps. We also fit a PCA model with $32$ components to the training data and run comparisons to our proposed generator. On the training set, PCA has a mean absolute error (MAE) of $0.054$, while our VAD-based solution achieves $0.036$ MAE. In \cref{fig:pca_vs_generative_qualitative_comparison}, we qualitatively compare the two solutions for sampling quality, noticing that our method produces more detailed and diverse samples. 

\begin{figure*}
\begin{center}
    \includegraphics[width=0.7\linewidth]{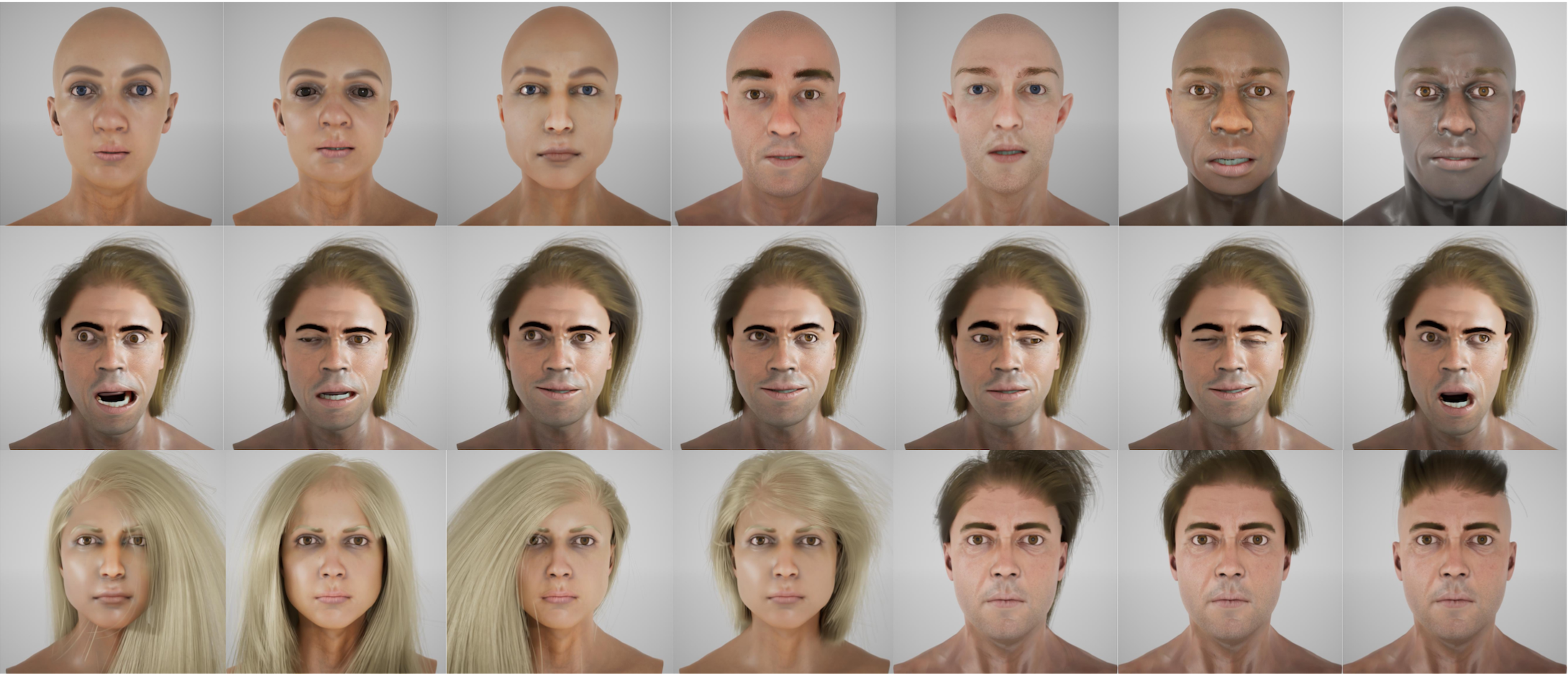}
\end{center}
\vspace{-2mm}
\caption{We show samples from the SPHEAR model that vary in shape and texture in the first row, facial expressions in the second row, and hair variations in the third row.}
\label{fig:variations}
\end{figure*}

\section{Additional Experiments}\label{sec:experiments}

In this section we introduce additional experiments for \textbf{registration}, and \textbf{3D} and \textbf{2D fitting}, further evaluating the accuracy of our proposed spherical registration as well as SPHEAR's accuracy, generalization capabilities, and fairness. 

\paragraph{Registration.} We compare our spherical embedding registration with other state-of-the-art registration methods in \cref{tbl:registration_experiments}. Our method achieves a $3\times$ improvement compared to \cite{Yoshiyasu2014}. We observed that both the deformation transfer \cite{sumner2004deformation} and the non-rigid ICP method from \cite{amberg2007optimal} are very sensitive to initialization. To achieve competitive results for the other methods, we first aligned the template scan using Procrustes analysis based on landmarks. For \cite{Yoshiyasu2014} we used our own implementation of ACAP, while for DT \cite{sumner2004deformation} and \cite{amberg2007optimal} we used the publicly available one \cite{trimesh}. 
\begin{table}[t]
\begin{center}
\small
\begin{tabular}[t]{l|rc}

Method  & Neutral (mm) & Expressions (mm) \\
\hline
DT \cite{sumner2004deformation} & 0.79 $\pm$ 0.23  & 0.90 $\pm$ 0.62 \\
NR-ICP \cite{amberg2007optimal} & 0.94 $\pm$ 0.24  & 1.17 $\pm$ 0.31 \\
ACAP \cite{Yoshiyasu2014} & 0.45 $\pm$ 0.15  & 0.51 $\pm$ 0.24 \\
\hline
\textbf{SR (Ours)} & \textbf{0.14 $\pm$ 0.03}  & \textbf{0.22 $\pm$ 0.15} 
\end{tabular}
\end{center}
\vspace{-3mm}
\caption{Registration experiments on \textbf{FHE3D-TEST}. We compare our spherical registration (SR) method with ACAP\cite{Yoshiyasu2014}, deformation transfer \cite{sumner2004deformation} (DT), and non-rigid ICP (NR-ICP) \cite{amberg2007optimal}. }
\label{tbl:registration_experiments}
\end{table}

\paragraph{3D fitting.}  We evaluate SPHEAR's generalization to unseen identities with different facial expressions by fitting it against test scans from \textbf{FHE3D-TEST}.
We compare SPHEAR on this task with state-of-the-art head models and report results in \cref{tbl:fitting_experiments}. For FLAME \cite{li2017learning} and Facescape \cite{yang2020facescape} we used the publicly available fitting code. For GHUM \cite{ghum2020} we used the standalone head model.
We used the same triangulated landmarks for all models obtained. We report average scan-to-mesh (S2M) errors of the face region on both scans in rest pose with neutral expression and for those featuring facial expressions.
SPHEAR is the most accurate model and is the most reliable (together with GHUM), with the lowest error variation.

\paragraph{2D fitting and fairness evaluation.} We evaluate SPHEAR's accuracy and diversity via fitting to 2D landmarks. 
For this task, we utilize a dataset containing 1.7k images of people evenly distributed across 17 geographical subregions.
Each image is annotated with 468 2D landmarks corresponding to the Facemesh \cite{grishchenko2020attention} semantics and with a skin tone using the Fitzpatrick scale \cite{fitzpatrickscale}. 
The 468 landmarks capture important variation in facial features and proportions across humanity.
To this end, we use those as a reference for evaluating the accuracy and fairness of our model.
We fit SPHEAR to each image using the 2D landmark reprojection error and analyze the remaining 2D  fitting error in pixels for each skin tone group separately, see \cref{tbl:skin_tone_landmarks_2d_experiments}.
Our model achieves the lowest errors across all head models and, more importantly, has consistent accuracy across all skin tones.  

\begin{table}[t]
\begin{center}
\small
\begin{tabular}{l|rc}
Method  & Neutral (mm) & Expressions (mm) \\
\hline
Facescape \cite{yang2020facescape} & 1.59 $\pm$ 0.66 & 1.60 $\pm$ 0.31 \\
FLAME \cite{li2017learning} & 1.58 $\pm$ 0.62  & 1.77 $\pm$ 0.39 \\
GHUM \cite{ghum2020} & 1.44 $\pm$ 0.24  & 1.63 $\pm$ 0.31 \\
\hline
\textbf{SPHEAR (Ours)} & \textbf{1.36 $\pm$ 0.28} & \textbf{1.58 $\pm$ 0.30}
\end{tabular}
\end{center}
\vspace{-3mm}
\caption{3D Fitting experiments on \textbf{FHE3D-TEST} for SPHEAR, GHUM\cite{ghum2020}, FLAME\cite{li2017learning} and Facescape\cite{yang2020facescape} for neutral and expression scans. Errors are expressed as the average distance from scan to mesh in millimeters. SPHEAR generalizes better than other head models to facial expressions of previously unseen identities.}
\label{tbl:fitting_experiments}
\end{table}

\begin{table}[t]
\begin{center}
\resizebox{\columnwidth}{!}{%
\small
\begin{tabular}[t]{l|cccccc}

Method  & ST1 \begin{minipage}{.02\textwidth}
      \includegraphics[width=1.\linewidth]{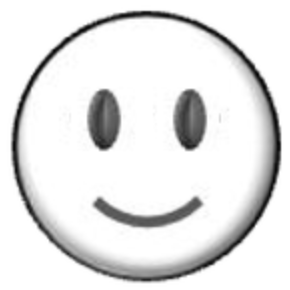}
    \end{minipage} & ST2 \begin{minipage}{.02\textwidth}
      \includegraphics[width=1.\linewidth]{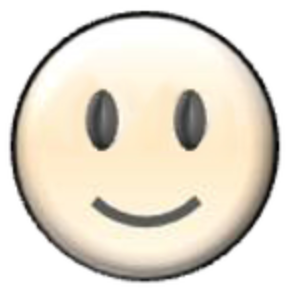}
    \end{minipage}& ST3 \begin{minipage}{.02\textwidth}
      \includegraphics[width=1.\linewidth]{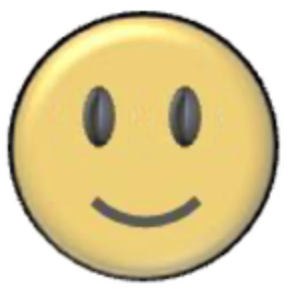}
    \end{minipage}& ST4 \begin{minipage}{.02\textwidth}
      \includegraphics[width=1.\linewidth]{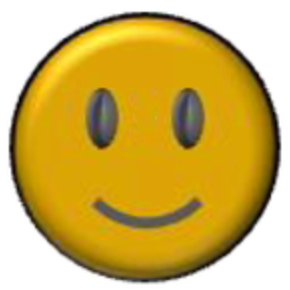}
    \end{minipage} & ST5 \begin{minipage}{.02\textwidth}
      \includegraphics[width=1.\linewidth]{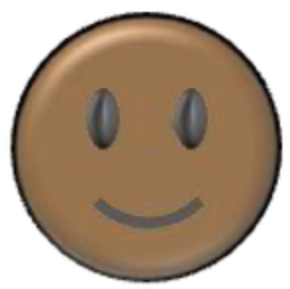}
    \end{minipage}& ST6 \begin{minipage}{.02\textwidth}
      \includegraphics[width=1.\linewidth]{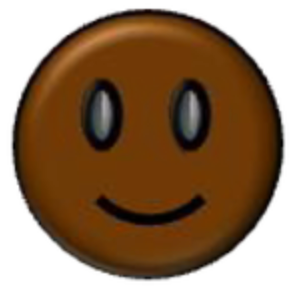}
    \end{minipage} \\
\hline
Facescape \cite{yang2020facescape} & 3.5 & 3.6 & 3.8 & 3.0 & 4.1 & 3.2\\
FLAME \cite{li2017learning} & 2.3 & 2.1 & 2.2 & 2.2 & 2.4 & 2.5 \\
GHUM \cite{ghum2020} & 2.2 & 2.0 & 2.1 & 2.1 & 2.3 & 2.5\\
\hline
\textbf{SPHEAR} & \textbf{1.7} & \textbf{1.7} & \textbf{1.7} & \textbf{1.7} & \textbf{1.7} & \textbf{1.8} \\

\end{tabular}%
}
\end{center}
\vspace{-4mm}
\caption{Skin-tone diversity analysis. We use a dataset of 1.7K images annotated with ground-truth 2D keypoints and 6 skin tone types (ST1 to ST6). Compared to other head reconstruction methods, SPHEAR achieves lower errors, consistently, across skin tones.}
\label{tbl:skin_tone_landmarks_2d_experiments}
\vspace{-4mm}
\end{table}

\section{Conclusions}

 To support a high degree of realism, we have developed a novel 3D registration methodology based on spherical embeddings, which benefits the construction of an accurate statistical 3D human head model. For completeness, we present additional components that enable fine-grained control in sampling diverse head shapes, facial expressions, skin appearance, and hair styles represented as strands. Experiments support the validity of design choices and the accuracy of our registration, reconstruction, and component generation techniques.
 
\noindent\textbf{Ethical Considerations.} Our methodology aims to decrease bias by supporting representation diversity and the generation of fair synthetic data. It facilitates bootstrapping for new domains and diverse subject distributions, where labeled data is often difficult to collect upfront. Our model is not intended or appropriate for any form of deep fakes.  To faithfully represent a person with our model, we would need access to high-quality 3D scans, which requires explicit user consent. Although our model can be animated and realistically rendered, there may still be a gap between our renderings and real video.

\noindent{\bf Limitations:} While our statistical model is the first of its kind, is still far from being perfect: we can do more in terms of modeling hair diversity (e.g. curliness), build more accurate normal/color reconstructions from raw scans, model micro-details, muscle and fat tissues.
Learning more detailed statistical components for partially hidden structures, e.g. teeth or the skull, may require additional imagery including CT scans.

\appendix

\section*{Supplementary Material}

In this supplementary material, we give examples of additional applications of our proposed spherical embedding, detail the hair and appearance networks, provide additional experiments, and showcase SPHEAR's diversity.

\section{Spherical geometry applications}
The spherical embedding we propose for registration has interesting applications beyond registration, as we show in the sequel. Concretely, we demonstrate how we can improve 3D landmark triangulation and how one can interpolate between two scans using the same methodology.

\subsection{Landmark triangulation} \label{sec:sphere_emb}
\label{sec:triangulation}
\paragraph{3D landmark triangulation.} One key prerequisite in learning $\mathcal{G}$ and $\mathcal{T}$ is the high fidelity registration of the template head mesh to the training set of scans. The first step in the registration process is to detect a set of semantically consistent 3D landmarks on each scan. For scalable, automatic and accurate head registration, we use FaceMesh~\cite{grishchenko2020attention} to obtain 478 2D landmarks, in 16 synthetic views $\Pb_{2d} \in \mathbb{R}^{16 \times  478 \times 2}$, with known camera parameters. We triangulate these points (through nonlinear optimization) to obtain the desired 3D landmarks.

\begin{figure}
    \centering
    \includegraphics[width=.9\linewidth]{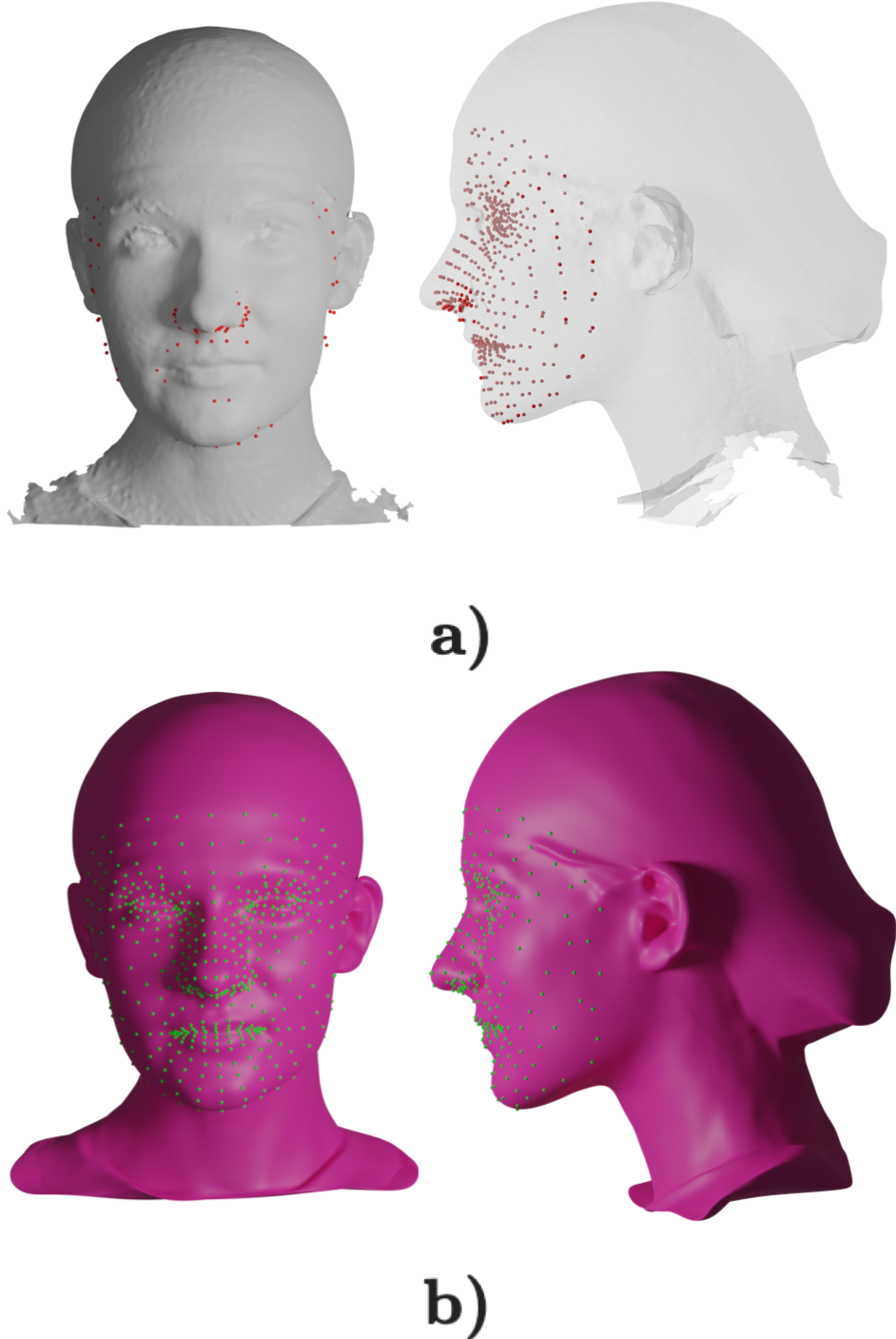}
    \caption{Landmark triangulation before and after manifold constraint.}
    \label{fig:triangulation}
\end{figure}

One issue with triangulated 3D landmarks from 2D detections is that they ignore the 3D geometry of the scan. Errors in 2D landmark inference will translate into errors in surface positioning. For example, the landmark for the tip of the nose might ``sink'' way below the nose surface, see \cref{fig:triangulation} a). We would like to constrain the 3D landmarks to remain on the surface during triangulation. We use our proposed spherical embedding, modeled as a trainable mapping with Siren activation networks \cite{sitzmann2020implicit}.

Similarly to Sec. 3.1 in the main paper, let $\xb \in \mathbb{R}^{1\times3}$ be a point on the scan surface, and let $\yb = f(\xb, \bphi_e) \in \mathbb{R}^{1\times3}$ be the point on the unit sphere, where $f$ is a trainable `encoder` network. To decode, we compute $\tilde{\xb} = g(\yb, \bphi_d) \in \mathbb{R}^{1\times3}$, where $g$ is the corresponding, trainable `decoder` network. Both networks are based on Siren activated MLPs, $S_e, S_d$
\begin{align}
    f(\xb, \bphi_e) &= \mathbf{\Pi}(\xb + S_e(\xb, \bphi_e)) \\
    g(\yb, \bphi_d) &= \yb + S_d(\yb, \bphi_d)
\end{align}
where $\bphi_e, \bphi_d$ are the trainable parameters of $S_e$ and $S_d$, respectively, and $\mathbf{\Pi}(\cdot)$ is the unit-sphere projection. In the following paragraph, we will drop the explicit dependency for clarity.

For a given scan $\mathbf{S} = \{\mathbf{V}_s, \mathbf{F}_s\}$ we sample $N_{s}$ points on its surface, $\Xb \in \mathbb{R}^{N_s\times3}$, and compute the reconstruction loss
\begin{align}
    \tilde{\Xb} &= g(f(\Xb)) \nonumber\\
    \mathcal{L}_{rec} &= \sum_{i} \|\Xb_i - \tilde{\Xb}_i\|_2.
\label{eq:tri_rec}
\end{align}

To ensure a smooth solution, we use $\mathcal{L}_{reg}$ as defined in Eq. 5, from the main paper.

The final loss is then $\mathcal{L}_{se} = \mathcal{L}_{rec} + w_{reg} \mathcal{L}_{reg}$, where $w_{reg}$ is the weight given to the regularization term, which depends on the global training step (it decreases in log space from $-1$ to $-7$).

After training the sphere embedding, we define a non-linear optimization problem that starts from the solution of the classical triangulation routine, $\Pb_0 \in \mathbb{R}^{478\times3}$, but in sphere space, \ie $\Qb_0 = f(\Pb_0)$, and uses the perspective operator $\mathbf{\Pi}_p$ which  projects the decoded landmarks $g(\Qb)$ in all camera views
\begin{align}
    \argmin_{\Qb} \|\mathbf{\Pi}_p(g(\Qb)) - \Pb_{2d}\|_2. 
\label{eq:tri_sphere}
\end{align}
After convergence, we retrieve the solution $\Pb = g(\Qb)$. This guarantees that reconstructed landmarks stay on the surface of the sphere and, implicitly, on the surface of the scan.

 In \cref{fig:triangulation} a), we present the originally triangulated points $\Pb_0$, from 16 different camera views. Notice how the points ``sink'' below the surface, failing to capture the nose and lips completely. Constraining the landmarks to sit on the surface of the scan, results in much better localization -- see \cref{fig:triangulation} b). The scan in \cref{fig:triangulation} b) is actually the result of the learned network $g$, when applied to an icosphere of depth 7, with $\approx 163k$ vertices and $\approx 330k$ triangular faces. The scan embedding error was, in this case, $0.11$ mm.

\paragraph{Architecture design choices.} We experimented with different options for our network architectures (i.e. $f, g, S_e, S_d$). We use a collection of scans $\{\Sb^i = \{\Vb^i_s, \Fb^i_s\}\}$, for which we compute their reconstruction error for all \textbf{vertices} as defined by $\mathcal{L}_{rec}(\Vb^i_s)$ in \cref{eq:tri_rec}. We also compute a per-vertex normal error, between the original vertex normals and the transformed vertex normals of $g(f(\Vb^i_s))$. In \cref{tbl:sphere_ablations}, we show ablation results. First ablation, ``w MLPs'', uses ReLU activated MLPs instead of Siren-based MLPs. The second one, ``w MLPs and RFE'', adds a Random Fourier Embedding \cite{tancik2020fourfeat} to the 3D points, before processing by the MLPs. This means constructing a kernel matrix $\mathbf{K} \in \mathbb{R}^{3 \times 64}$, with entries drawn from $\mathcal{N}(0, \sigma)$. The scale $\sigma$ is assigned through validation. The RFE embeddings will be $[\sin(\xb\mathbf{K}), \cos(\xb\mathbf{K})] \in \mathbb{R}^{128}$. Both architectures performed worse than our model, in terms of both errors, while also displaying some artifacts.

\begin{table}[t]
\small
\begin{center}
\resizebox{\linewidth}{!}{%
\begin{tabular}[t]{l||c|c}
Method  & Reconstr.\ Error (mm)  & Normal Error (\textdegree)\\
\hline
w MLPs & $0.37$ & $6.31$ \\
w MLPs and RFE & $0.25$ & $4.51$ \\
\hline
\textbf{Ours} & $\mathbf{0.15}$ & $\mathbf{2.88}$\\
\end{tabular}
}
\end{center}
\caption{Ablations of different architectural choices for the sphere mapping network. We show how replacing the Siren-based MLPs with ReLU activated MLPs -- with and without Random Fourier Embedding (RFE) -- leads to much higher errors.}
\label{tbl:sphere_ablations}
\end{table}
\subsection{Registration}

\paragraph{Implementation details.} We set the modulation codes, $\cb_s, \cb_t$, to be 3-dimensional, and we initialize them from a truncated standard normal distribution. We did not observe improved performance with a larger code size. For the Siren networks and Siren modulators, we use 5 intermediate layers with 128 hidden neurons, for a total of $\approx 66$k and $\approx 68$k parameters, respectively. Note that we change the intermediate activation to \texttt{tanh} for the Siren modulators, as \texttt{ReLU} was failing to converge. The networks are trained for 10000 steps, with 512 sampled points each iteration (on the scan surfaces), and 256 sampled points on the sphere. The whole training takes roughly $3$ minutes on a Nvidia Titan XP graphics card. The main bottleneck is the second-order Hessian computation (through which we have to pass gradients), for the regularizer $\mathcal{L}_{reg}$.

\section{Hair network}
\paragraph{Texture space embedding.} We embed each hair strand $\hh \in \mathbb{R}^{100 \times 3}$, where $100$ is the number of control points of the curve, in the UV texture space of the template scalp. To facilitate training, we also parameterize in the TBN (tangent, bitangent and normal) space, such that the variance across directions is reduced \cite{rosu2022neuralstrands}. We also observed that by encoding them by relative offsets improves performance -- i.e. $\hh^{r} = \{\hh_{i+1} - \hh_{i}\}_{i\leq 99} \in \mathbb{R}^{99 \times 3}$.

\paragraph{Legendre polynomial bases.} We observe that training a network directly for generating control points, leads to very spiky, non-smooth results. To address this issue, we turn towards expressing the hair strands as parameterized curves. We choose to use Legendre polynomials. They form a system of orthogonal polynomials $P_{n\geq0}(\mathbf{x})$, defined over the interval $[-1, 1]$, such that:

\begin{align}
    \int_{-1}^{-1} P_n(\xb) P_m(\xb) d\xb = 0, \forall n \neq m
\end{align}

We fit a 5-th degree polynomial with Legendre bases (which we call \textit{Legendre polynomial}), i.e. $L_5(\xb) = \sum_{0\leq n \leq 5} w_n P_{n}(\xb)$, to all hair strands, for each Cartesian component $x, y, z$ separately. The weights $w$ will uniquely describe each hair strand, and we get $\hh_l^r \in \mathbb{R}^{3\times 6}$. To make sure that distances between polynomials make sense, we have to ensure that the orthogonal polynomial bases $P_n$ are also ortho-normal. For that, we compute the norm of each of basis and divide by it:
\begin{align}
    \|P_n\| = \sqrt{\int_{-1}^{1} P^2_n(\xb) d\xb}
\end{align}

Note that only by having Legendre polynomials (or other described by ortho-normal bases) ensures that distances between curves are proportional to distances between samples on the curves.

\paragraph{Limitations.} Note that a polynomial with degree $n$ can have at most $n-1$ turning points. This is a limiting factor in modeling more complex hair strands, with a lot of \textit{curliness}. We chose the degree to $5$, as the dataset on which we train does not exhibit curly hair. However, we can control curliness after generation, with some post-processing, if needed.

\begin{figure*}[t!]
    \centering
    \includegraphics[width=1.\linewidth]{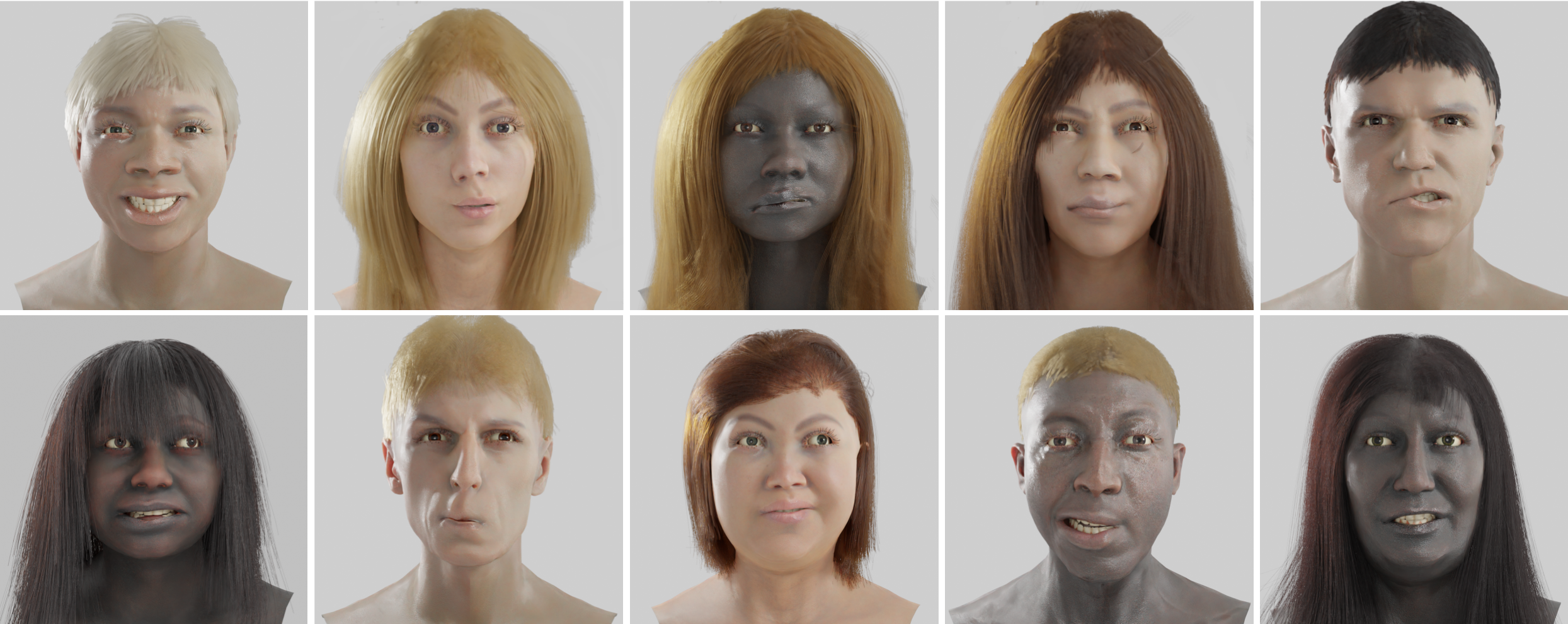}
    \caption{Samples from our full SPHEAR generative model varying in identity (including appearance), expression, eye gaze, and hair style.} 
    \label{fig:full_sampling}
\end{figure*}

\paragraph{Network details.} We take a variational auto-decoder approach \cite{zadeh2019variational, hao2020dualsdf}, and assign a unique identifier to each training example. The network, similar to \cite{karras2017progressive}, goes from a code an initial map ($8\times8\times128$)
, and to a final map $256\times256\times18$ through sequential deconvolutions and upsampling blocks. Like in \cite{karras2019style}, we also replace the nearest-neighbour upsampling with bilinear sampling. The complete network has $1,718,476$ trainable parameters and is trained with \texttt{Adam} optimizer \cite{kingma2014adam}, batch size of $16$, learning rate of $1e-3$ and a decay of $0.99$.

\paragraph{Intersection-free penalty.} Once trained, samples might, qualitatively, look smooth and plausible, but they will not necessarily respect physical constraints: i.e. they might intersect with the head template. In order to address this, we revisit the training procedure. We embed the head template into a sphere (similar to \ref{sec:sphere_emb}), and re-attach the hair training samples. We re-parameterize the strands $\mathbf{h}$ in terms of $(1 + s_i)[x_i, y_i, z_i]$, where $x_i^2 + y_i^2 + z_i^2 = 1$ and $s_i \geq 0$., where $i$ is the control point. This is basically expressing any control point in terms of spherical coordinates plus a non-negative scaling factor, which guarantees that any control point is not inside the sphere, and, implicitly, not inside the head.

\begin{figure}[t!]
    \centering
    \includegraphics[width=.8\linewidth]{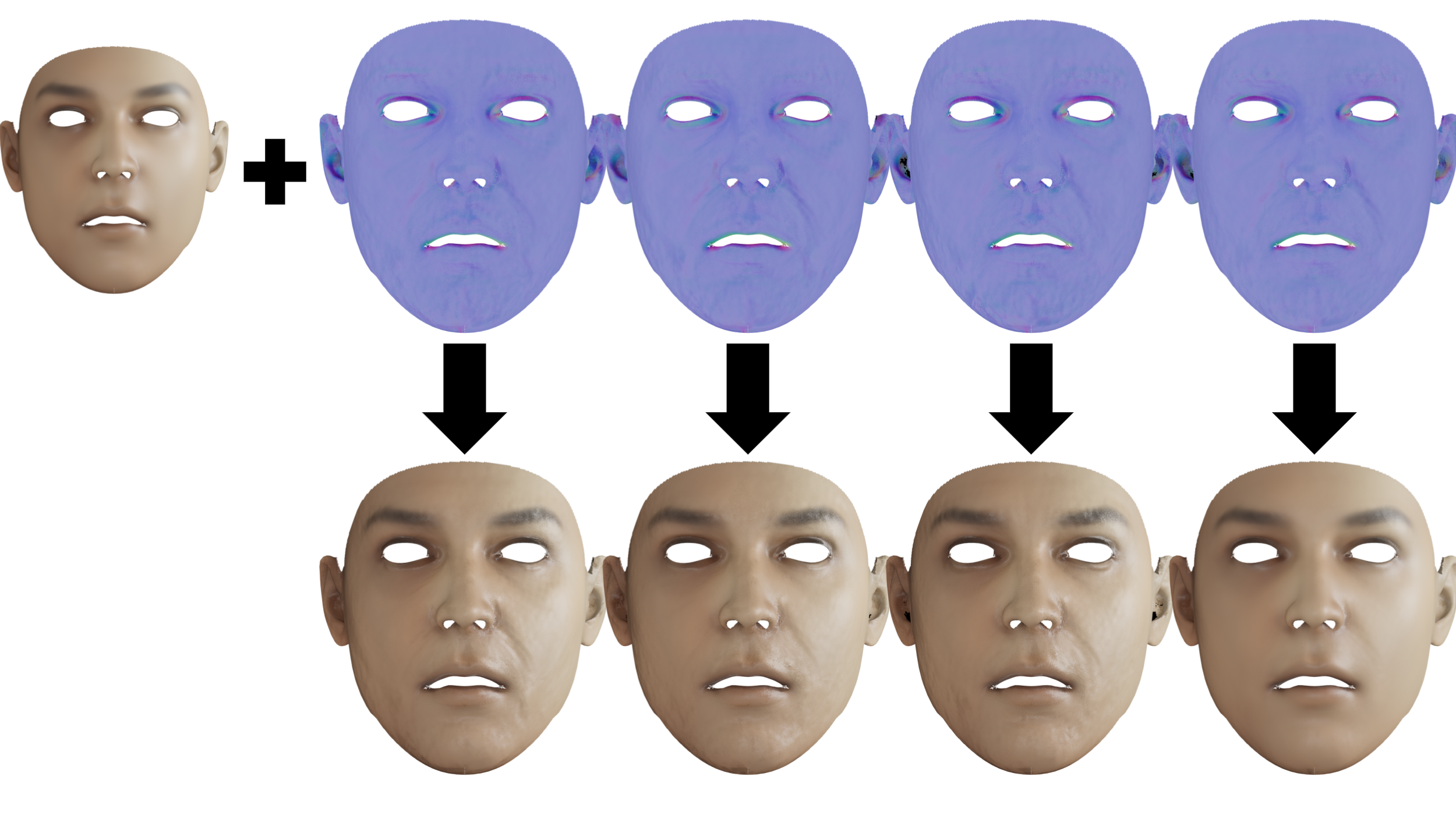}
    \caption{Sampled normals from our VAD. \textbf{Top left.} A constant albedo map. \textbf{Top right.} 4 sampled normals. \textbf{Bottom right.} The result of shading by combining normals and albedo. Notice the differences in skin appearance, showing wrinkles and skin imperfections. \textbf{Best viewed digitally, zoomed in.}}
    \label{fig:normals}
\end{figure}

\section{Appearance network}

\paragraph{Normal generation.} We also show the quality of sampled normals, from our appearance generator $\mathcal{T}$, in \cref{fig:normals}. Normals are important for increasing the visual fidelity of the renderings, and it is a cheap way of encoding skin imperfections, wrinkles or geometric deformations.

\paragraph{Network details.} We use a very similar architecture to the hair network, but we make changes to the number of processing layers, as the final map is $4$ times as big, from $256\times 256$ to $1024 \times 1024$. Note that in practice, we have 2 generative appearance models, one for albedo and for normals. Each of them has roughly $1, 516, 000$ trainable parameters.

\section{Diversity}
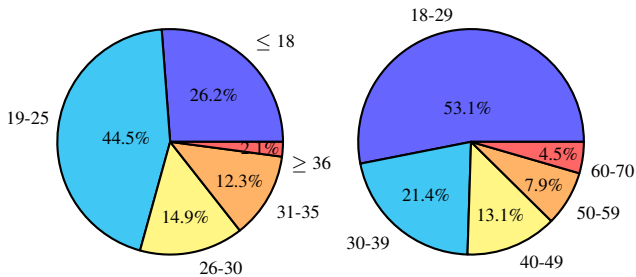
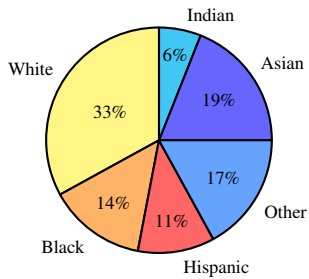
\begin{figure}
\scriptsize

\begin{subfigure}[b]{1.\linewidth}
\begin{tikzpicture}

\pie[xshift=0cm,scale=0.5]%
  {26.2/$\leq$ 18, 44.5/19-25, 14.9/26-30, 12.3/31-35, 2.1/$\geq$ 36}

\pie[xshift=4cm,scale=0.5]{53.1/18-29, 21.4/30-39, 13.1/40-49, 7.9/50-59, 4.5/60-70}
\end{tikzpicture}
\caption{Body mass index (\textbf{left}) and age (\textbf{right}) distributions.}
\vspace{10mm}
\end{subfigure}
\vspace{5mm}
\begin{subfigure}[b]{1.\linewidth}
\centering
\begin{tikzpicture}
\pie[xshift=0cm,scale=0.5]{19/Asian, 6/Indian, 33/White, 14/Black, 11/Hispanic, 17/Other}
\end{tikzpicture}
\caption{Ethnicity/skin tones distributions.}
\end{subfigure}
\vspace{-4mm}
\caption{Diversity distribution for the \textbf{FHE3D} dataset. We show distributions over three criteria: age, body mass index (BMI) and ethnicity/skin tone.}
\label{fig:Diversity}
\end{figure}

To show that our samples lead to fair machine learning algorithms, we report the distributions of various attributes for the scan data used in training our generators.
We consider the skin color distribution of our captured subjects in our \textbf{FHE3D} dataset, covering different ethnicities, the body mass index and the age (see  \cref{fig:Diversity}). As it can be observed, the distribution is fairly weighted across all considered brackets.Once trained, SPHEAR allows us to sample diverse 3D heads of people with varying identity, expression, eye gaze, and hair style, see \cref{fig:full_sampling}.

\begin{figure}[H]
    \centering
    \includegraphics[width=1.0\linewidth]{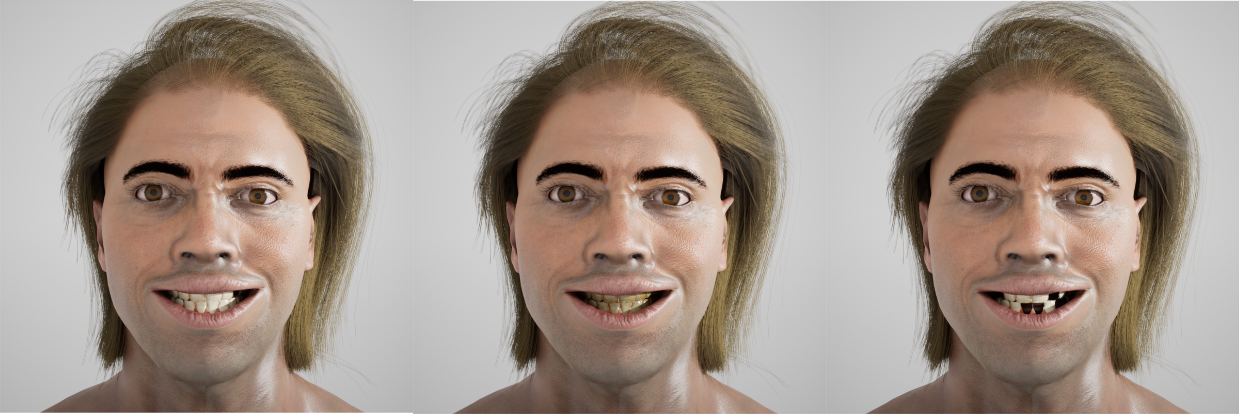}
    \caption{Sample teeth texture variation. Best seen in color.} 
    \label{fig:teeth_texture_variation}
    \vspace{3mm}
\end{figure}

\paragraph{Teeth Texture Variation.}
Capturing teeth geometry for teeth is difficult given the current systems. We rely instead on a procedural method to control erosion effects, teeth discoloration, presence of cavities and lack of teeth. We implement the effects as a material shader compatible with existing photorealistic rendering engines. We illustrate such examples in \cref{fig:teeth_texture_variation}.

\section{Applications}

In the sequel we demonstrate three different applications of SPHEAR. First, we represent 3D head scans with SPHEAR through fitting. Next, we provide additional results for the 2D fitting experiment from the main paper. Finally, we use SPHEAR for synthetic data generation for two different tasks: 2D landmark detection and dense semantic labeling. Using SPHEAR, obtaining accurate labels is cheap and we show that trained predictors generalize well to real images.

\paragraph{3D Fitting.} Given a textured 3D scan geometry $\mathbf{S}$ of the head of a person, we are interested in fitting the parameters of our geometric component $\mathcal{G}$,  $\mathbf{\Phi}=\left(\bbeta^{s}, \bbeta^{e}, \btheta\right)$. The SPHEAR mesh vertex positions are then given by this geometric component $\mathbf{V_{\mathbf{\Phi}}}=\mathcal{G}(\mathbf{\Phi})$. Similar to the registration pipeline, we use FaceMesh \cite{grishchenko2020attention} to obtain $478$ 2D landmarks, in 16 different rendered views $\Pb_{2d} \in \mathbb{R}^{16 \times  478 \times 2}$ of the scan. We triangulate these points (through nonlinear optimization) and refine them using the sphere embedding approach (\cf \cref{sec:triangulation}). We denote the final 3D landmarks on the scan surface as $\Pb^s$. We then optimize the head model parameters with respect to (1) the alignment to the 3D landmarks, together with (2) an implicit ICP-type loss driven by the learned spherical embedding. The optimization loss can be written as:
\begin{equation}
    \argmin_{\mathbf{\Phi}} \lambda_{lmks}\|W_{lmks}\mathbf{V}_{\mathbf{\Phi}} - \Pb^s\|_2 + \lambda_{rec}\|g(f(\mathbf{V}_{\mathbf{\Phi}})) - \mathbf{V}_{\mathbf{\Phi}}\|_2 \nonumber
\end{equation}
where $W_{lmks} \in  \mathbb{R}^{478 \times 12201}$ is a sparse matrix used to regress the landmarks locations from the SPHEAR vertices and $\lambda_{\{lmks, rec\}}$ are scalars used to weigh the two terms. In the second term, $g(f(\mathbf{V}_{\mathbf{\Phi}}))$ represents the reconstruction of the vertices through the sphere embedding of the \textbf{scan}. This guarantees that these reconstructed points are \textbf{on} the scan surface. This acts similar in spirit to ICP, where one instead of using $g(f(\mathbf{V}_{\mathbf{\Phi}}))$ would consider the nearest-neighbors of $\mathbf{V}_{\mathbf{\Phi}}$ on the scan. Compared to classical ICP, where the process of finding the nearest point at each optimization step is non-differentiable, our objective is fully differentiable, is faster to compute and leads to smaller errors (mesh-to-scan average distances, in mm): $2.1$ vs $3.0$ on neutral shapes, and $2.87$ vs $3.7$ on shapes with expressions.

\begin{figure*}
    \vspace{5cm}
    \centering
    \includegraphics[width=1.0\linewidth]{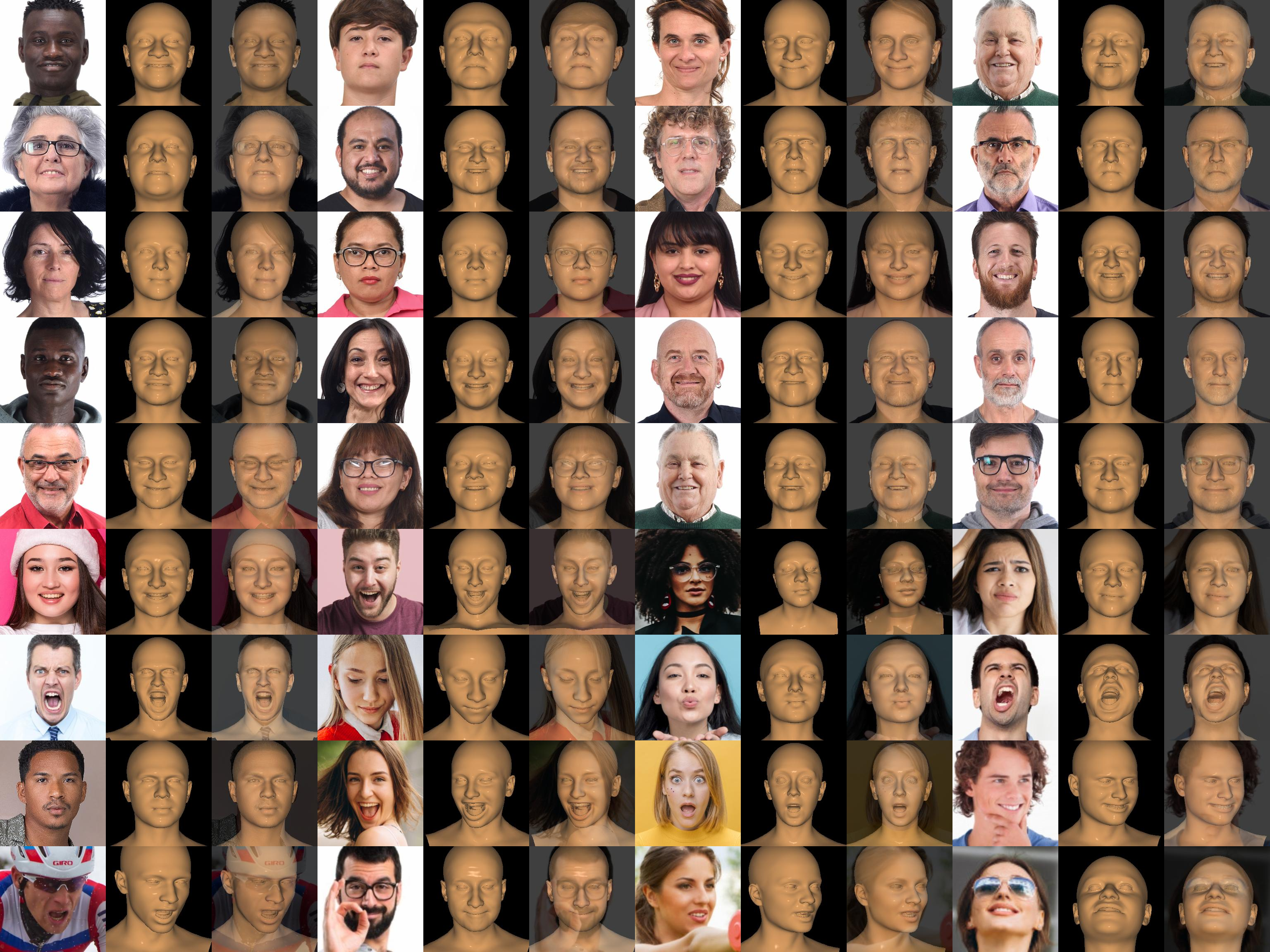}
    \caption{Sample results on in-the-wild images for 2D landmark fitting. For each input image we show the SPHEAR fitting alone and as an overlay. Note how SPHEAR accurately models the expressions and detailed facial features. Best seen in color and zoomed in.} 
    \label{fig:sample_2d_fitting}
    \vspace{5cm}
\end{figure*}

\paragraph{2D Fitting.} We provide more examples of our 2D fitting results in figure \cref{fig:sample_2d_fitting}. As illustrated by the experiments in the main paper our model fits well to diverse identities, facial expressions and head poses. 

\begin{figure}
\begin{center}
    \includegraphics[width=1.0\linewidth]{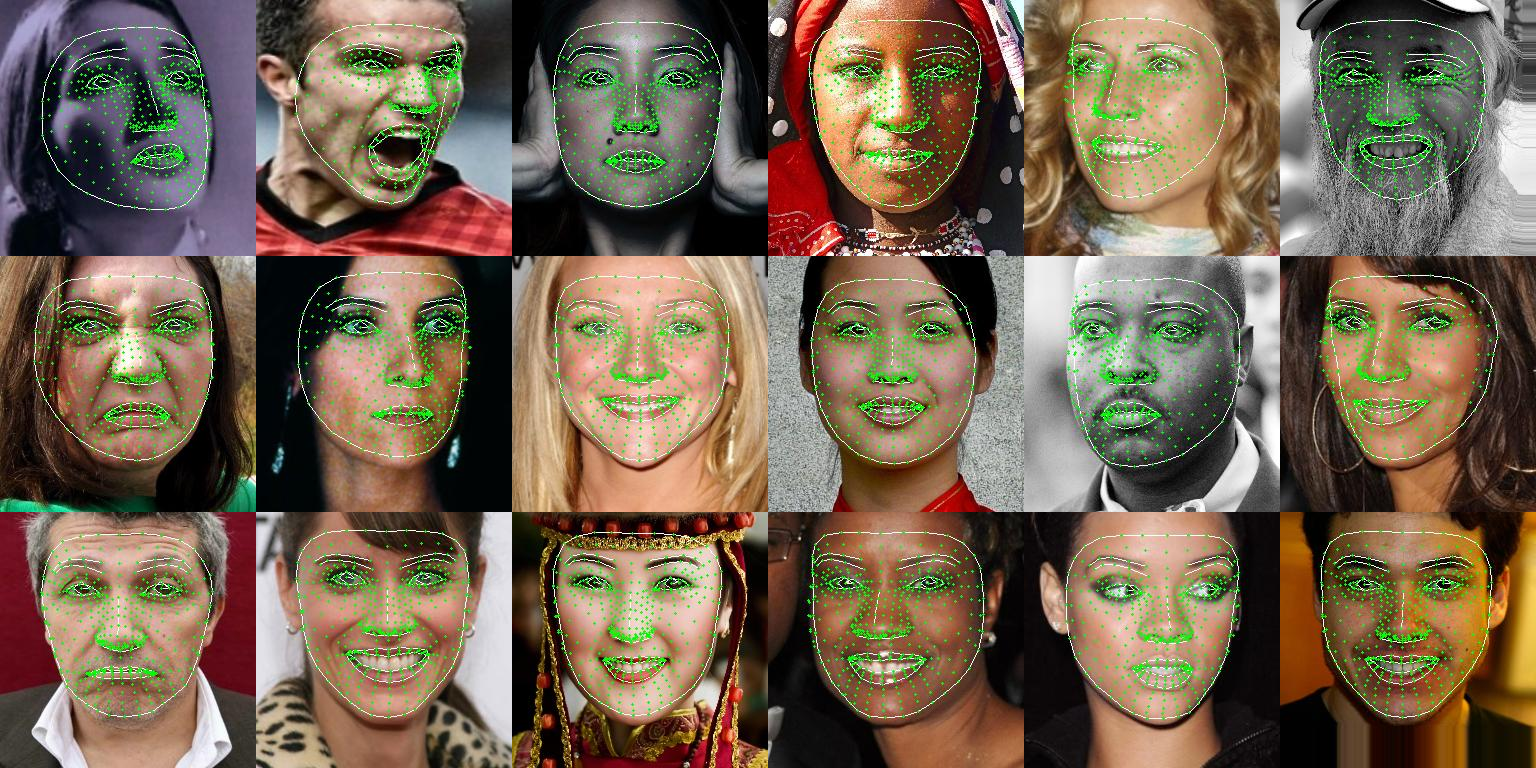}
\end{center}
\vspace{-2mm}
\caption{We use SPHEAR to generate a synthetic dataset of human faces that is used to train a dense landmark prediction model \cite{grishchenko2020attention} (478 landmarks, including eye gaze). We illustrate its output on sample images from the test set of \textbf{300W} \cite{sagonas2016300} that contain diverse identities, different illuminations, and different facial expressions.}
\label{fig:applications-landmarks-examples}
\vspace{1cm}
    \centering
    \includegraphics[width=1.0\linewidth]{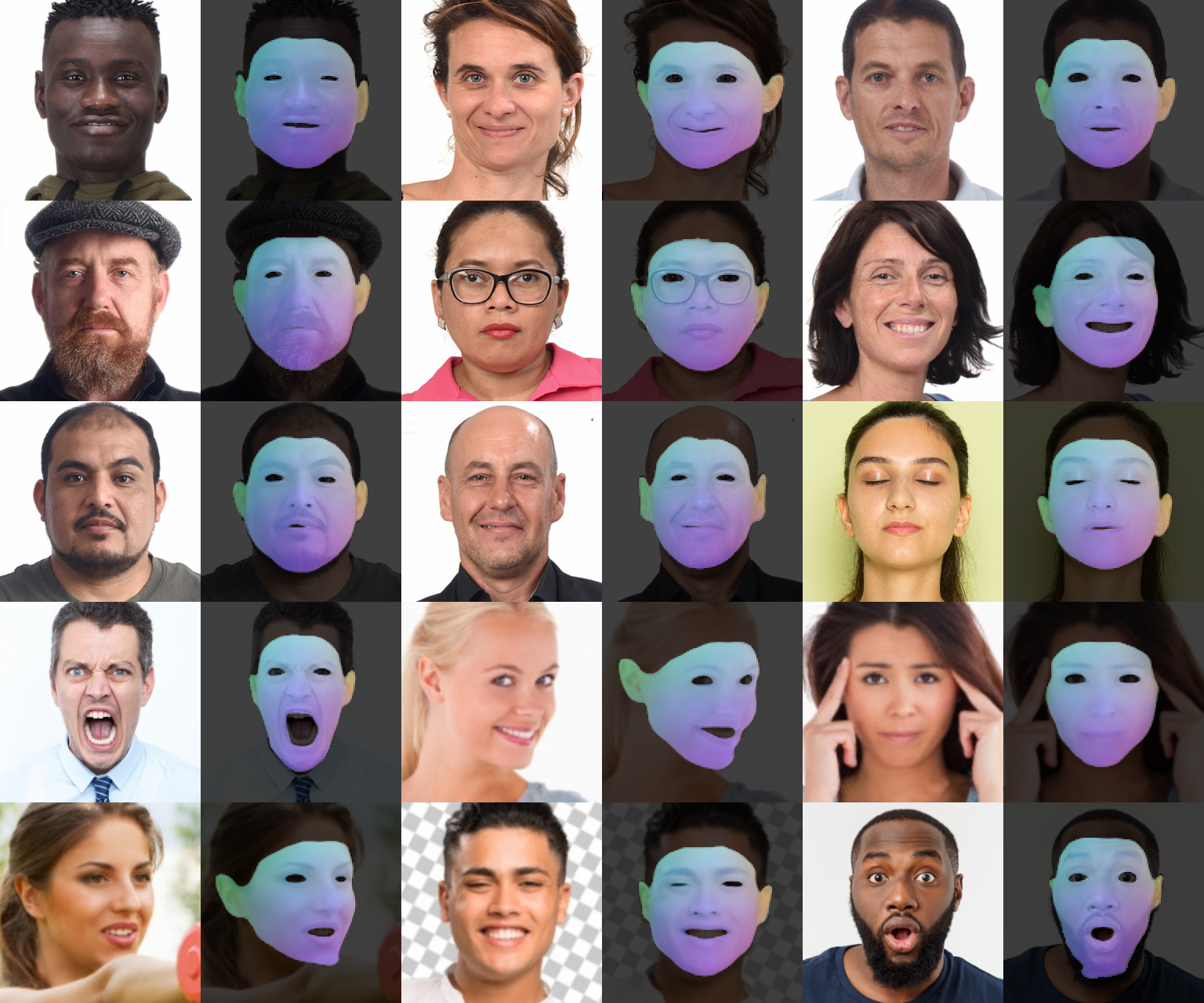}
    \caption{Sample results on in-the-wild images for our dense prediction network trained on synthetic SPHEAR data. Predicted mask color encodes positions on the SPHEAR template.} 
    \label{fig:dense_reconstructions}
\end{figure}

\begin{table}
\begin{center}
\begin{tabular}[t]{l||r|r|r}
Method  & Common & Challenging & Private \\
\hline
Real & 3.33 & 5.59 & 4.53 \\
Synthetic & 3.52 & 6.62 & 4.91 \\

\end{tabular}
\end{center}
\vspace{-2mm}
\caption{Landmark localization errors on \textbf{300W} \cite{sagonas2016300}. We report separate numbers for each subset of the test data. Expressed as normalized pixel error normalized by inter-ocular distance.}
\label{tbl:landmarks_2d_68}
\end{table}

\paragraph{Landmark Predictions.} We evaluate the usefulness of our SPHEAR model in a 2D landmark detection task. For this, we generate a synthetic dataset of rendered faces from various viewpoints, with sampled attributes from $\{\mathcal{G}, \mathcal{H}, \mathcal{T}\}$, a discrete set of eye colors and HDRI backgrounds. In total, we generate 60K samples for training and 10K samples for validation using 500 HDRI images from the public domain \cite{polyhaven}. These allow us to vary illumination patterns and backgrounds. We trained a lightweight real-time production-level face mesh network (1M parameters) \cite{grishchenko2020attention} to predict 478 dense landmarks using purely synthetic supervision.  In \cref{fig:applications-landmarks-examples}, we show that our trained model generalizes to unseen images in-the-wild. For quantitative evaluation, we trained a similar real-time network on an in-house dataset, consisting of 20K real images with manually annotated landmarks. We test both methods on the public \textbf{300W} dataset \cite{sagonas2016300}. For both networks, we also trained an additional two-layer domain adaptation network to regress the $68$ landmarks that are annotated in \textbf{300W}. We report results in \cref{tbl:landmarks_2d_68} and observe on par performance. Overall, this reduces the need for costly annotations, as is often the case when using manually labeled data, while maintaining comparable performance.

\paragraph{Dense Predictions.} Besides the landmarks estimation application we also explored a dense semantic prediction task. For the each synthetic data example generated using SPHEAR, we rendered a semantic mask of the ground truth head geometry containing the 3D template coordinates as attributes. We also considered rasterizing the UV coordinate attributes instead, however the UV coordinates are not continuous especially close to the texture seams\cite{bhatnagar2020loopreg, alldieck2021imghum}. We then trained a UNet\cite{ronneberger2015u} network $\mathbf{q}$, taking as input RGB images, in order to predict rasterized semantic masks. For each coordinate $p \in \mathbf{q}\left( I \right)$ in the output mask, the dense correspondence network predicts a 3D coordinate $\mathbf{x}_{p}$ on the SPHEAR template. We present some sample qualitative results in \cref{fig:dense_reconstructions}. 

\vspace{1cm}

{\small
\bibliographystyle{ieeenat_fullname}
\bibliography{egbib}
}

\end{document}